\documentclass[journal]{IEEEtran}

\usepackage{cite}
\usepackage{color, soul}
\soulregister\cite7
\soulregister\ref7
\usepackage[mathscr]{eucal}

\usepackage[pdftex]{graphicx}
\usepackage{subfig}
\graphicspath{ {pics/} }
\DeclareGraphicsExtensions{.png}

\usepackage{multirow}

\usepackage[cmex10]{amsmath}
\DeclareMathOperator*{\argmax}{argmax}
\usepackage{cleveref}

\usepackage{array}

\usepackage{url}

\begin{document}
	\title{\bf A General Arbitration Model for Robust Human-Robot Shared Control with Multi-Source Uncertainty Modeling}

	\author{Songpo~Li, Michael Bowman, and  
		Xiaoli~Zhang$^{*}$,~\IEEEmembership{Member,~IEEE}, 
		\thanks{Songpo Li is a Postdoctoral Associate in the Department of Electrical and Computer Engineering at Duke University, Durham, NC 27708 USA (e-mail: songpo.li@duke.edu).}
		\thanks{Michael Bowman is a Ph.D. Candidate in the Department of Mechanical Engineering at Colorado School of Mines, Golden, CO 80401 USA (e-mail:mibowman@mines.edu).}
		\thanks{Xiaoli Zhang is an Associate Professor in the Department of Mechanical Engineering at Colorado School of Mines, Golden, CO 80401 USA ($^{*}$corresponding author, phone: 303-384-2343; fax: 303-273-3602; email: xlzhang@mines.edu).}
		
		}
	\maketitle
	
	\begin{abstract}
		Shared control in teleoperation leverages both human and robot's strengths and has demonstrated great advantages of reducing the difficulties in teleoperating a robot and increasing the task performance. One fundamental question in shared control is how to effectively allocate the control power to the human and robot. Researchers have been subjectively defining the arbitrate policies following conflicting principles, which resulted in great inconsistency in the policies. We attribute this inconsistency to the inconsiderateness of the multi-resource uncertainty in the human-robot system. To fill the gap, we developed a multi-source uncertainty model that was applicable to various types of uncertainty in real world, and then a general arbitration model was developed to comprehensively fuse the uncertainty and regulate the arbitration weight assigned to the robotic agent. Beside traditional macro performance metrics, we introduced objective and quantitative metrics of robotic helpfulness and friendliness that evaluated the assistive robot's cooperation at micro and macro levels. Results from simulations and experiments showed the new arbitration model was more effective and friendly over the existing policies and was robust to coping with multi-source uncertainty. With this new arbitration model, we expect the increased adoption of human-robot shared control in practical and complex teleoperation tasks.
	\end{abstract}


\section{Introduction}
\IEEEPARstart{T}{eleoperating} a robot allows operators to carry out tasks remotely with the robot as a medium while viewing its live video feedback. This indirect interaction brings in many advantages including increased motion precision and strength, and remote access to work fields that might be inaccessible or hazardous to the operator \cite{taylor2016medical, artemiadis2010emg, kofman2005teleoperation}. However, successfully teleoperating the robot for a task is often difficult and complex due to indirect and often oriented visualization, indirect manipulation with the robot, and physical discrepancies between a human hand and robot hand \cite{rybarczyk2002contribution, healey2008speculation}. Due to those difficulties, the operator easily feels lost in the visual feedback of the work field and has a difficult time figuring out how to operate the control interface to achieve desired robot motions. In the meantime, the above control process often associates with high mental and physical workload. Ways that can reduce the teleoperation complexity are being actively investigated \cite{okamura2004methods, livatino2014stereoscopic, zhai2016adaptive}.

Increasing robots' intelligence and autonomy level to allow them to generate (semi-)autonomous behaviors and proactively assist in achieving the operator's goal has demonstrated great advantages \cite{dragan2013policy, khademian2011dual, li2015continuous}. This shared control promotes the role of the robot from a passive motion follower or executor to a collaborative partner that shares in the control of the physical components of the system. Shared control leverages both strengths of the human's adaptability for decision making in dynamic,  uncertain environments and the robot's automation capability for accomplishing a task faster, easier, and decreasing the physical and mental demands on the human \cite{muelling2015autonomy,javdani2015shared,crandall2002characterizing}. 

Shared control allocates the relative amount of control power between the operator and robot based on a predefined arbitration policy, and definition of that policy has always been one of the fundamental problems. A proper arbitration policy gives proper amount of control power to each party at the correct time to maximize their advantages and minimize their disadvantages. Due to the lack of theoretical support, most arbitration policies have been artificially defined by researchers based on subjective intuition and have resulted in great varieties, including conflicting policies and results. 

Here, we attribute the conflicted policy principles and policy varieties to the lack of comprehensive consideration of the multi-source uncertainty in the (semi)autonomous robotic system. Two main types of uncertainty are uncertainty in human intent understanding and uncertainty in automation execution. The first type is a result of the ambiguous human motion, cluttered environment, and imperfect intent inference algorithm; while, the second type is a combination of the sensing uncertainty, control uncertainty, and hardware uncertainty. The existence of this uncertainty that is not considered can result in misestimating the autonomous system's capability of providing effective assistance, leading to inappropriate control allocation between the human and the robot and then task failures, performance decrease, and human resistance. 

The effective shared control requires to allocate the appropriate amount of control power to the human and robot corresponding to the various uncertainty conditions. For practical deployment of effective shared control, we model the multi-source uncertainty of various types and levels and investigate how those types of uncertainty influence the allocation process. Particularly, our major contributions are:

\begin{enumerate}
\item A general uncertainty model that models the multi-source uncertainty at various magnitude in a shared-control robotic system.
\item A general arbitration model that copes with multi-source uncertainty and leverages the strength of human operators and robotic agents.
\item Two objective and quantitative metrics that evaluate a robotic agent's helpfulness and friendliness at both micro and macro levels during human-robot shared control. 
\end{enumerate}

\section{Related Work}

Shared-control lies between manual control from human operators and fully autonomous control by an intelligent-sufficient robotic agent \cite{schilling2016towards}. Its introduction and popularity are results of the great need to assist the human operator and facilitate the complexity and difficulty in manually teleoperating a robotic platform. Effectively allocating the control power between the operator and robotic agent has always been a key research question. Many researchers have followed a principle that assigns more control power to the robot when it is closer to the target \cite{dragan2013policy, anderson2010semi,sellner2005user}. This policy suggests that while the robot gets closer to the target, the likelihood of the robot approaches the correct target increases. In this case, the researchers believe more control power given to the robot can release the control workload of the operator. In addition, the major reason for this principle is the uncertainty of the user's approaching intent in the unstructured task. We refer the policies that follow this principle as positive policies as the robotic agent's control power is positively related to the decreased distance to the inferred target. This principle has resulted in various policy profiles by adjusting the power increase ratio, inflection points, or minimum and maximum control power \cite{loizou2007mixed, weber2009position, javdani2018shared}. Moreover, conflicting research results have been reported by various researchers about performance and preference \cite{marayong2002effect, you2012assisted, kim2011autonomy}.

In contrast to these positive policies, some researchers have argued that the robotic agent's control power should decrease since the unavoidable uncertainty in automation execution as a combination of sensing uncertainty, algorithm uncertainty, and hardware uncertainty \cite{owan2015uncertainty}. This principle questions the robotic system's accuracy and suggests increased failure when the robot is close to the target. We refer a policy that follows this principle as a negative policy, as the robot's control power is negatively related to the decreased distance to the target. Little work has been reported about this negative policy, and the uncertainty modeling and arbitration modeling are still open questions. 

Another open problem with those positive and negative policies is that they have not considered the various uncertainty levels. Different human intent inference mechanisms in different work environments and different robotic systems could result in various levels of uncertainty. The current work attempts to fit all uncertainty conditions with the same arbitration policy, which certainly results in unideal performance outcomes. 

Besides the great inconsistencies of arbitration policies, the subjectively defined arbitration policies have not been well evaluated. After the implementation of an arbitration policy, measures, such as the task success rate, task completion time, and subjective surveys were only available to evaluate that policy \cite{reddy2018shared, giesbrecht2017safeguarding, nikolaidis2017human}. Those metrics measure a policy's overall performance at the macro level. As stated earlier that an appropriate arbitration policy gives appropriate amount of control power to each party at the correct time, those macro metrics cannot discover how a policy performs in dynamic. As a result, after evaluations of the policies, the researcher cannot quantitatively explain how such a performance is achieved and what take-aways are delightful for other researchers while they are defining their arbitration policy. Moreover, the success rate and completion time are performance-orientated instead of evaluating how well the robotic agent cooperates with the human operator. Novel metrics that can quantitatively explain how a robotic agent cooperates with the human operator at the micro level would be beneficial in studying the shared control policies.

\section{Method}

This section introduces the formulation of the shared-control model in detail. There are three main modules, 
\begin{enumerate}
\item human intent inference for the approaching target,
\item uncertainty modeling in the intent inference and robotic autonomy, and
\item formulation of the shared-control model. 
\end{enumerate}

\subsection{Complementary Intent Inference}

\begin{figure}[tb]
\centerline{\includegraphics[width=0.8\linewidth]{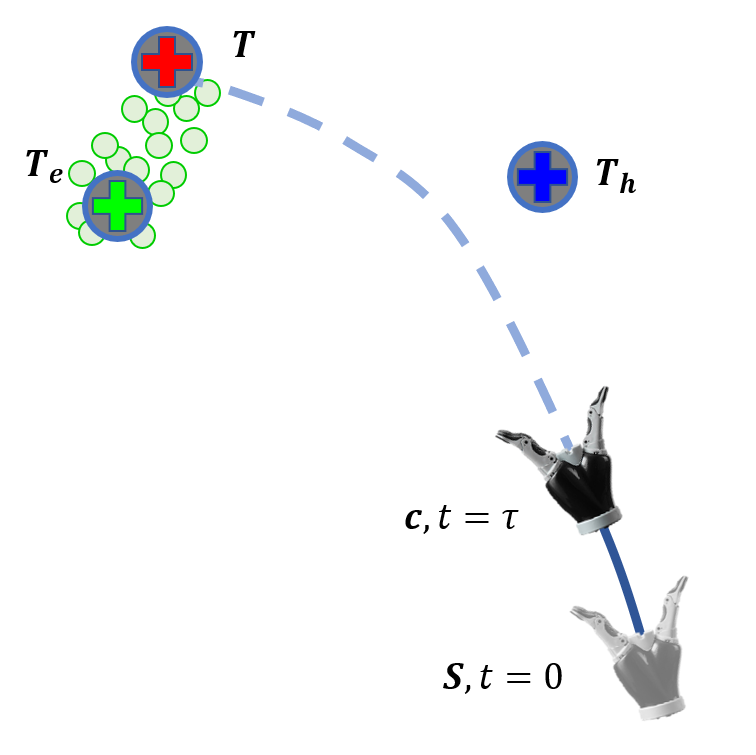}}
\caption{Demonstration of fusing the eye-hand modalities for human intent inference. Three big circles with a cross inside are possible targets. Small green dots are measured gaze points, and the blue line (solid and dashed) is the end effector's motion trajectory. The robotic agent could easily mistake $T_h$ as the target if using the hand information only, and mistake $T_e$ as the target using the eye information. Fusing two modalities enables the robotic agent to avoid this mistake completely or realize and correct this mistake as early as possible.}
\label{fig1}
\end{figure}

Knowing the human intent is the prerequisite for providing timely and appreciate assistance. A multimodal intent inference method is developed here based on the natural eye-hand cooperation, as human eyes lead hands to fall on the manipulation target during the natural human manipulation. Thus, this multimodal method takes advantage of the complementary spatial-temporal information of the eye and hand modalities. Eye information gives the robotic agent earlier access to the manipulation target, while it could be unstable in spatial due to the nature of eye movements. The hand motion can strongly imply whether an object is the manipulation target, while it has considerable temporal lag and could be fuzzy when passing by an irrelevant object. Fusing two modalities allows the robotic agent to realize the human intent earlier with high confidence. Fig. \ref{fig1} demonstrates the advantage of fusion of the eye-hand modalities. If using a single modality for the intent inference, the robotic agent may easily mistake $T_e$ or $T_h$ as the target. Fusing two modalities enables the robotic agent to avoid this mistake completely or realize and correct this mistake as early as possible.

\subsubsection{Intent Inference based on the Eye Modality}

We formulate the human intent inference using the eye modality  \eqref{IIEM} similar to our previous work \cite{li2014implicit, li2013attention, li2015attention}, and $\mathcal{T}$ is a set of accessible objects that could be the target and $E$ is a sequence of eye-gaze data measured since eyes dwelling. In other words, the robotic agent infers the human intent $T_e$ that maximizes the posterior probability while knowing a sequence of eye-gaze data. The eye-gaze data includes eye dwelling time, gaze speed, pupil dilation, and gaze concentration. The utilization of multiple eye-gaze features is to reduce the influence of visual distractions and thus to improve the accuracy. The eye-gaze data will also pass through a sliding window filter to remove high-frequency involuntary eye movements. 
\begin{equation}
T_e=\argmax_{T\in\mathcal{T}}p(T|E)
\label{IIEM}
\end{equation}

\subsubsection{Intent Inference based on the Hand Modality}

The human intent inference from the hand modality employs a trajectory-based inference method \cite{dragan2013policy,gopinath2016human}. As shown in Fig. \ref{fig1}, given the historical trajectory $L_{S\xrightarrow{}c}$, from the starting point $S$ to the current location $c$, the robotic agent maximizes the following posterior probability \eqref{IIHM}.
The probability $P(T|L_{S\xrightarrow{}c})$ uses the principle of maximum entropy as the formula, where the probability of an object as the intent exponentially decreases as the cost of the approaching it increases with the given trajectory. 
\begin{equation}
T_h=\argmax_{T\in\mathcal{T}}p(T|L_{S\xrightarrow{}c})
\label{IIHM}
\end{equation}

\subsubsection{Inference Fusion}

The fusion of two modalities is through a Bayesian inference approach  \eqref{IF} \cite{castanedo2013review}. Given a sequence of eye-gaze data and the historical approaching trajectory, the intent inference is to maximize the fused posterior probability. The inferred target is annotated as the nominal target $T_n$. This nominal target could be the same as the true target, but also could be different as this complimentary inference method or any other inference is to reduce the change of mistaking a wrong target to its best but cannot be able to eliminate this mistake. 
\begin{equation}
T_n=\argmax_{T\in\mathcal{T}}p(T|E)p(T|L_{S\xrightarrow{}c})
\label{IF}
\end{equation}

\subsection{Uncertainty Modeling}

\subsubsection{Confidence under Intent Inference Uncertainty}

Due to the uncertainty introduced by the clustered environment, human's ambiguous motion, and imperfect inference algorithm, it is inevitable to mistake a close-by object as the target. The uncertainty from various sources propagates for individual modalities. We model the propagated uncertainty of the intent inference using eye and hand modalities each as a three-dimension (3D) Gaussian distribution \eqref{CIIU1}-\eqref{CIIU2}, and it describes the true target distributes around the inferred target following a Gaussian distribution. This Gaussian distribution has the inferred target as the mean, and the variance is known by evaluating a modality's historic inference performance, which is mean of squared deviations from the inferred targets to the true targets. In the equations, $\Sigma_e$ and $\Sigma_h$ are variances of eye and hand modalities respectively, and $\sigma_{e,i}^2$ and $\sigma_{h,i}^2$, $i\in\{x,y,z\}$ are eye and hand modalities' variances in three axes, respectively.  We assume there is no correlation between any two axes in the distribution for problem simplification. The intent inferred through fusing the eye-hand modalities consequently has a Gaussian uncertainty distribution \eqref{CIIU3} with $T_n$ as the mean and $\sigma_{eh,i}^2$, $i\in\{x, y, z\}$ as the variances on three axes \cite{bromiley2003products}. $\sigma_{eh,i}^2$, $i\in\{x, y, z\}$ are computed as \eqref{CIIU4}. The scaling factor $\mathcal{S}$ is a constant computed upon $T_e$, $T_h$, $\Sigma_e$ and $\Sigma_h$.

\begin{multline}
f_{e}(T)=p\left(T_{e}, \Sigma_{e}\right) =\frac{1}{(2 \pi)^{\frac{3}{2}} \sigma_{e, x} \sigma_{e, y} \sigma_{e, z}} \bullet \\ \exp \left[-\frac{\left(T_{x}-T_{e, x}\right)^{2}}{2 \sigma_{e, x}^{2}}-\frac{\left(T_{y}-T_{e, y}\right)^{2}}{2 \sigma_{e, y}^{2}}-\frac{\left(T_{z}-T_{e, z}\right)^{2}}{2 \sigma_{e, z}^{2}}\right]
\label{CIIU1}
\end{multline}

\begin{multline}
f_{h}(T)=p\left(T_{h}, \Sigma_{h}\right) =\frac{1}{(2 \pi)^{\frac{3}{2}} \sigma_{h, x} \sigma_{h, y} \sigma_{h, z}} \bullet \\ \exp \left[-\frac{\left(T_{x}-T_{h, 1}\right)^{2}}{2 \sigma_{h, x}^{2}}-\frac{\left(T_{y}-T_{h, 2}\right)^{2}}{2 \sigma_{h, y}^{2}}-\frac{\left(T_{z}-T_{h, 3}\right)^{2}}{2 \sigma_{h, z}^{2}}\right]
\label{CIIU2}
\end{multline}

\begin{multline}
f_{n}(T)=p\left(T_{n}, \Sigma_{e}\right) \cdot p\left(T_{n}, \Sigma_{h}\right) \\ =\mathcal{S} \frac{1}{(2 \pi)^{\frac{3}{2}} \sigma_{e h, x} \sigma_{e h, y} \sigma_{e h, z}} \bullet\\  \exp \left[-\frac{\left(T_{x}-T_{n,1}\right)^{2}}{2 \sigma_{e h, x}^{2}}-\frac{\left(T_{y}-T_{n, 2}\right)^{2}}{2 \sigma_{e h, y}^{2}}-\frac{\left(T_{z}-T_{n,3}\right)^{2}}{2 \sigma_{e h, z}^{2}}\right]
\label{CIIU3}
\end{multline}

\begin{equation}
\sigma_{e h, i}^{2}=\frac{\sigma_{e, i}^{2} \sigma_{h, i}^{2}}{\sigma_{e, i}^{2}+\sigma_{h, i}^{2}}
\label{CIIU4}
\end{equation}

We assume the uncertainty of the inferred intent has the same variance, $\sigma_n^2$ on three axes, and this can be achieved by re-evaluating the deviation between the inferred intent and the true intent on three axes. Rewriting the uncertainty distribution function as \eqref{CIIU5}, and thus the probability of one object being the true target only depends on its distance to the inferred target and the pre-known distribution variance. 
\begin{multline}
f_{n}(T)= \mathcal{S} \frac{1}{(2 \pi)^{\frac{3}{2}} \sigma_{n}^{3}} \bullet \\ \exp \left[-\frac{\left(T_{x}-T_{n, x}\right)^{2}+\left(T_{y}-T_{n y}\right)^{2}+\left(T_{z}-T_{n z}\right)^{2}}{2 \sigma_{n}^{3}}\right]\\= \mathcal{S} \frac{1}{(2 \pi)^{\frac{3}{2}} \sigma_{n}^{3}} \exp \left[-\frac{d^{2}}{2 \sigma_{n}^{2}}\right]\label{CIIU5}
\end{multline}

\begin{figure}[tb]
\centerline{\includegraphics[width=0.9\linewidth]{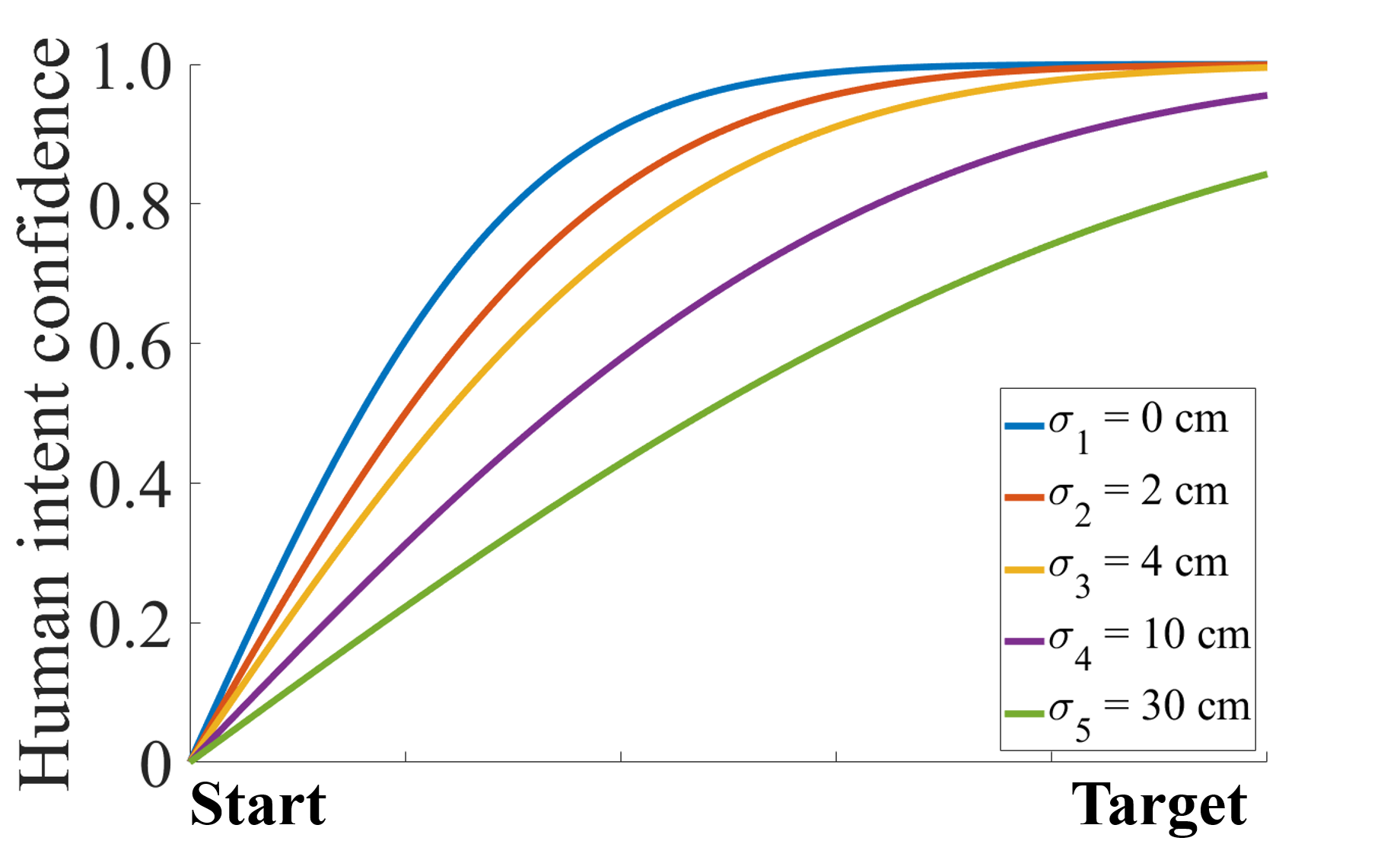}}
\caption{Demonstration of human intent confidence when the intent inference has various levels of uncertainty, $\sigma_n$.}
\label{fig2}
\end{figure}

Following the same principle that the closer the inferred target gets the more likely it is the true target, we define that intent confidence as a function of the distance to the inferred target and regulated by the uncertainty variance \eqref{CIIU6}-\eqref{CIIU7}. $D$ is a constant threshold that defines the function range of shared control. Outside this range, there is too much uncertainty in the system and the robotic agent does not contribute to the control of the robot; within this range, the robotic agent shares the control with the human operator following the defined arbitration function. $\aleph$ is a carefully defined regulation function of the $\sigma_n$ and $D$, and $E$ is a constant. Fig. \ref{fig2} are samples of the confidence in human intent with various uncertainty levels, and the confidence gradually increases while the end effector approaching the target. 

\begin{equation}
\operatorname{conf_{in}}(d)
\begin{cases}
\text{erf} \left(\frac{1}{a} \left( 1 - \frac{d}{D}\right)\right) & \text{if $d \leq D$} \\
0 & \text{if $d>D$}
\end{cases}
\label{CIIU6}
\end{equation}

\begin{equation}
a=1-\frac{\left(\sigma_{n}-D\right)^{2}}{\mathrm{E}}\label{CIIU7}
\end{equation}

\subsubsection{Confidence under Autonomy Uncertainty}

Intent uncertainty associates with the problem “which target”, while the autonomy uncertainty attempts to solve the problem “where the robot should approach to reach that target.”  This ubiquitous uncertainty could be mainly from the sensing accuracy of the target location and hardware limitation or misalignment for reaching that location. Due to this uncertainty, the robotic agent faces potential failures to handle the task independently, and with higher uncertainty the lower the robotic agent's confidence in handling the task. In addition, even though with the same level of uncertainty, the failure chance increases, and the confidence lowers when the end effector approaches the target. Those changes highlight the characteristics of a robotic agent's confidence in handling the task independently. 

To mathematically model this correlation between the autonomy uncertainty and the robotic agent's confidence, we assume the sensing uncertainty $f_s(T_s)$, and the hardware uncertainty $f_w(P)$, follow two 3D Gaussian distributions $\mathcal{N}(T,\Sigma_s)$ and $\mathcal{N}(T_s,\Sigma_w)$, respectively \eqref{CAU1}-\eqref{CAU2}. These two types of uncertainty can be estimated through a trial and error method. $\mathcal{N}(T,\Sigma_s)$ represents the distribution probability of target measure $T_s$ when the true target is at $T$, and $\Sigma_s$ is the distribution's variance matrix. $\mathcal{N}(T_s,\Sigma_w)$ represents the distribution probability of the end effector's final location $P$, when it attempts to approach $T_s$, and $\Sigma_w$ is the distribution's variance matrix. We also assume there is no correlation between any two axes in either distribution for problem simplification.  

\begin{multline}
f_{s}\left(T_{s}\right)=\mathcal{N}\left(T, \Sigma_{s}\right)=\frac{1}{(2 \pi)^{\frac{3}{2}} \sigma_{s, x} \sigma_{s, y} \sigma_{s, z}} \bullet \\\exp \left[-\frac{\left(T_{s, x}-T_{x}\right)^{2}}{2 \sigma_{s, x}^{2}}-\frac{\left(T_{s, y}-T_{y}\right)^{2}}{2 \sigma_{s, y}^{2}}-\frac{\left(T_{s, z}-T_{z}\right)^{2}}{2 \sigma_{s, z}^{2}}\right]
\label{CAU1}
\end{multline}

\begin{multline}
f_{w}(P)=\mathcal{N}\left(T_{s}, \Sigma_{w}\right)=\frac{1}{(2 \pi)^{\frac{3}{2}} \sigma_{w, x} \sigma_{w, y} \sigma_{w, z}} \bullet \\\exp \left[-\frac{\left(P_{x}-T_{s, x}\right)^{2}}{2 \sigma_{w, x}^{2}}-\frac{\left(P_{y}-T_{s, y}\right)^{2}}{2 \sigma_{w, y}^{2}}-\frac{\left(P_{z}-T_{s, z}\right)^{2}}{2 \sigma_{w, z}^{2}}\right]
\label{CAU2}
\end{multline}

The distribution between $P$ and $T$ is \eqref{CAU3}, and this is the integral of the joint distribution of $f_s(T)$ and $f_w(P)$. As there is no correlation between any two axes in $\mathcal{N}(T,\Sigma_s)$ or $\mathcal{N}(T_s,\Sigma_w)$, the integral can be calculated along each axis separately \eqref{CAU4}–\eqref{CAU8}. Thus, $P$ also follows a Gaussian distribution with $T$ as the mean and $\Sigma_s+\Sigma_w$ as the variance. 

\begin{multline}
f(P)=\iiint_{-\infty}^{+\infty} p\left(P | T_{s}\right) p\left(T_{s}\right) d T_{s}\\=\iiint_{-\infty}^{+\infty} f_{s}\left(T_{s}\right) f_{w}(P) d T_{s}
\label{CAU3}
\end{multline}

\begin{multline}
f(P)=\prod_{i \in\{x, y, z\}} \int_{-\infty}^{+\infty} \frac{1}{2 \pi \sigma_{s, i} \sigma_{w, i}} \bullet \\ \exp \left[-\frac{\left(T_{s, i}-T_{i}\right)^{2}}{2 \sigma_{s, i}^{2}}-\frac{\left(P_{i}-T_{s,i}\right)^{2}}{2 \sigma_{w, i}^{2}}\right] d T_{s, i}
\label{CAU4}
\end{multline}

\begin{multline}
f(P)=\prod_{i \in\{x, y, z\}} \frac{1}{\sqrt{2 \pi\left(\sigma_{s, i}^{2}+\sigma_{w, i}^{2}\right)}} \exp \left[-\frac{\left(P_{i}-T_{i}\right)^{2}}{2\left(\sigma_{s, i}^{2}+\sigma_{w, i}^{2}\right)}\right] \\ \bullet \int_{-\infty}^{+\infty} \frac{1}{\sqrt{2 \pi} \sigma_{s w, i}} \exp \left[-\frac{\left(T_{s, i}-T P_{i}\right)^{2}}{2 \sigma_{s w, i}^{2}}\right] d T_{s, i}
\label{CAU5}
\end{multline}

\begin{equation}
\sigma_{sw, i}^{2}=\frac{\sigma_{s, i}^{2} \sigma_{w, i}^{2}}{\sigma_{s, i}^{2}+\sigma_{w, i}^{2}}
\label{CAU6}
\end{equation}

\begin{equation}
T P_{i}=\frac{T_{s, i} \sigma_{w, i}^{2}+T_{s, i} \sigma_{s, i}^{2}}{\sigma_{s, i}^{2}+\sigma_{w, i}^{2}}
\label{CAU7}
\end{equation}

\begin{multline}
f(P)=\prod_{i \in\{x, y, z\}} \frac{1}{\sqrt{2 \pi\left(\sigma_{s, i}^{2}+\sigma_{w, i}^{2}\right)}} \bullet\\
\exp \left[-\frac{\left(P_{i}-T_{i}\right)^{2}}{2\left(\sigma_{s, i}^{2}+\sigma_{w, i}^{2}\right)}\right]
\label{CAU8}
\end{multline}

While the end effector approaches the target in an arbitrary manner, the probability of encountering the target at a distance $R$ can be related to the cumulative distribution function. This is the integral in infinite space $V$, with a spherical hollow of radius $R$. For simplification, we can re-evaluate the variances so that three axes have the same variance $\sigma_a^2$. Thus, the encountering probability can be computed as \eqref{CAU9} and \eqref{CAU10}. $p_c(R)$ has a reversed sigmoid shape and is regulated by $\sigma_a^2$. In contrast, the failure probability of approaching can then be computed as $1-p_c(R)$.

\begin{multline}
p_{c}(R)=\int_{V} f(P) d V= \\ \frac{1}{(2 \pi)^{\frac{3}{2}} \sigma_{a}^{3}} \int_{0}^{2 \pi} d \theta \int_{0}^{\pi} \sin \varphi d \varphi \int_{R}^{+\infty} \rho^{2} \exp \left(-\frac{\rho^{2}}{2 \sigma_{a}^{2}}\right) d \rho
\label{CAU9}
\end{multline}

\begin{multline}
p_{c}(R)= 1-\operatorname{erf}\left(\frac{R}{\sqrt{2 \sigma_{a}^{2}}}\right)+\sqrt{\frac{2}{\pi \sigma_{a}^{2}}} R \exp \left(-\frac{R^{2}}{2 \sigma_{a}^{2}}\right)
\label{CAU10}
\end{multline}

\begin{figure}[tb]
\centerline{\includegraphics[width=0.9\linewidth]{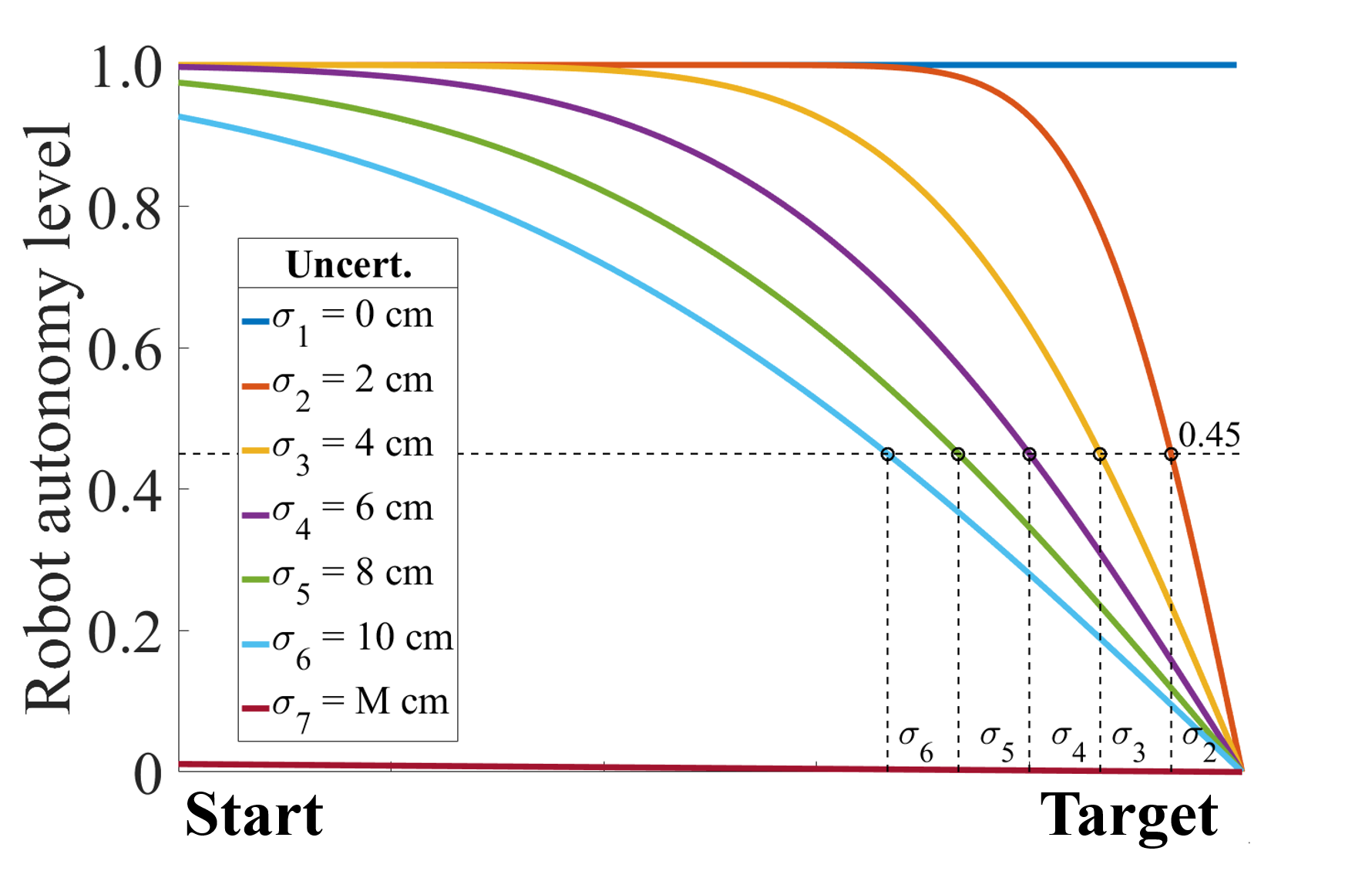}}
\caption{Demonstration of robot autonomy confidence when autonomy has various levels of uncertainty. According to the definition, when the end effector is within a distance $\sigma_a$ to the target the autonomy confidence falls below $\Gamma$. $\Gamma$ is 0.45 in this demonstration. M indicates an extremely large value.}
\label{fig3}
\end{figure}

We relate the robotic agent's confidence in its autonomy to its failure probability, and we redefine this failure probability as \eqref{CAU11} and \eqref{CAU12}. The $D$ is still the function range of the shared control, and $\Gamma$ is a constant selected to regulate the confidence decrease behavior. Fig. \ref{fig3} demonstrates robotic confidence in autonomy as a function of the distance to the target. This new definition cannot only simplify the relationship while holding a certain degree of similarity to the original definition in \eqref{CAU10} but also offers better controllability with the help of $\Gamma$. Involvement of $b$ constraints the autonomy confidence always under $\Gamma$ when the end effector is within a distance $\sigma_a$ to the target. For example, when the end effector is less than 2 cm away from the target the autonomy confidence is less than 0.45 and gradually reduces to zero. 

\begin{equation}
\operatorname{conf_{au}}(d)=
\begin{cases}
\operatorname{erf} \left(\frac{d}{b} \right) & \text{if $d \leq D$} \\
0 & \text{if $d>D$}
\end{cases}
\label{CAU11}
\end{equation}

\begin{equation}
b=\frac{\sigma_{a}}{D \cdot \operatorname{erfinv(\Gamma)}}
\label{CAU12}
\end{equation}

\subsection{Formulation of Shared Autonomy}

The arbitration weight of the robotic agent can then be defined as \eqref{FSA1} as a combination of the confidence in the intent inference and robotic autonomy. $\alpha$ is a function of the distance to the inferred target and also regulated by the level of uncertainty in the gaze modality, hand modality, sensing, and robot hardware. The final motion command $\boldsymbol{m}_t$ sent to the robot's end-effector is a combination of the human motion input $\boldsymbol{x}_t$, and the robotic agent's motion input $\boldsymbol{y}_t$ \eqref{FSA2}.

\begin{equation}
\alpha =\operatorname{conf_{in}}(d) \cdot \operatorname{conf_{au}}(d)
\label{FSA1}
\end{equation}
\begin{equation}
\boldsymbol{m}_{t}=(1-\alpha) \boldsymbol{x}_{t}+\alpha \boldsymbol{y}_{t}
\label{FSA2}
\end{equation}

\begin{figure}[t]
	\centerline{\includegraphics[width=\linewidth]{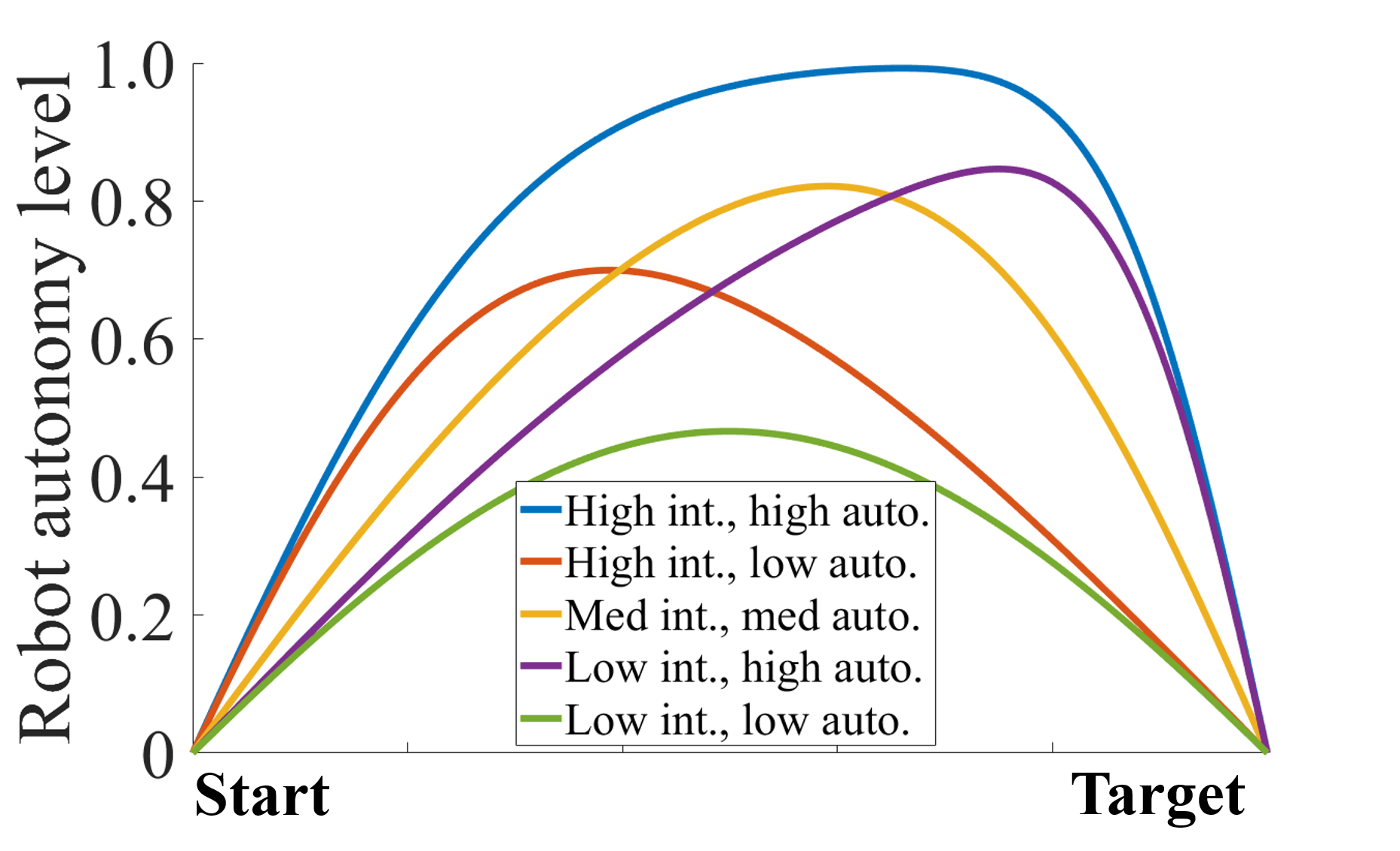}}
	\caption{Demonstration of effects of various levels of confidence in intent inference (int.) and in autonomy (auto.) on the arbitration in shared control. If both confidences are low, weight assigned to the robotic agent will follow the green line. If both confidence are high, the blue line will be followed. }
	\label{fig4}
\end{figure}

Fig. \ref{fig4} illustrates the effects of two types of confidence, $\operatorname{conf}_{\text {in }}$ and $\operatorname{conf}_{\text {au}}$, on the arbitration weight assigned to the robotic agent. If the human operator's weight is higher than the robotic agent's, the human operator is dominant (likely region close to the start point). When both types of confidence are high, the robotic agent becomes dominant earlier and contributes more to the motion of the end effector. When either type of confidence is low, the robotic agent contributes less. We name this new arbitration model as bell-shaped policy due to the policy profile and comparison with the positive and negative policies.

\subsection{Shared-Autonomy Framework}

\begin{figure}[tb]
\centerline{\includegraphics[width=\linewidth]{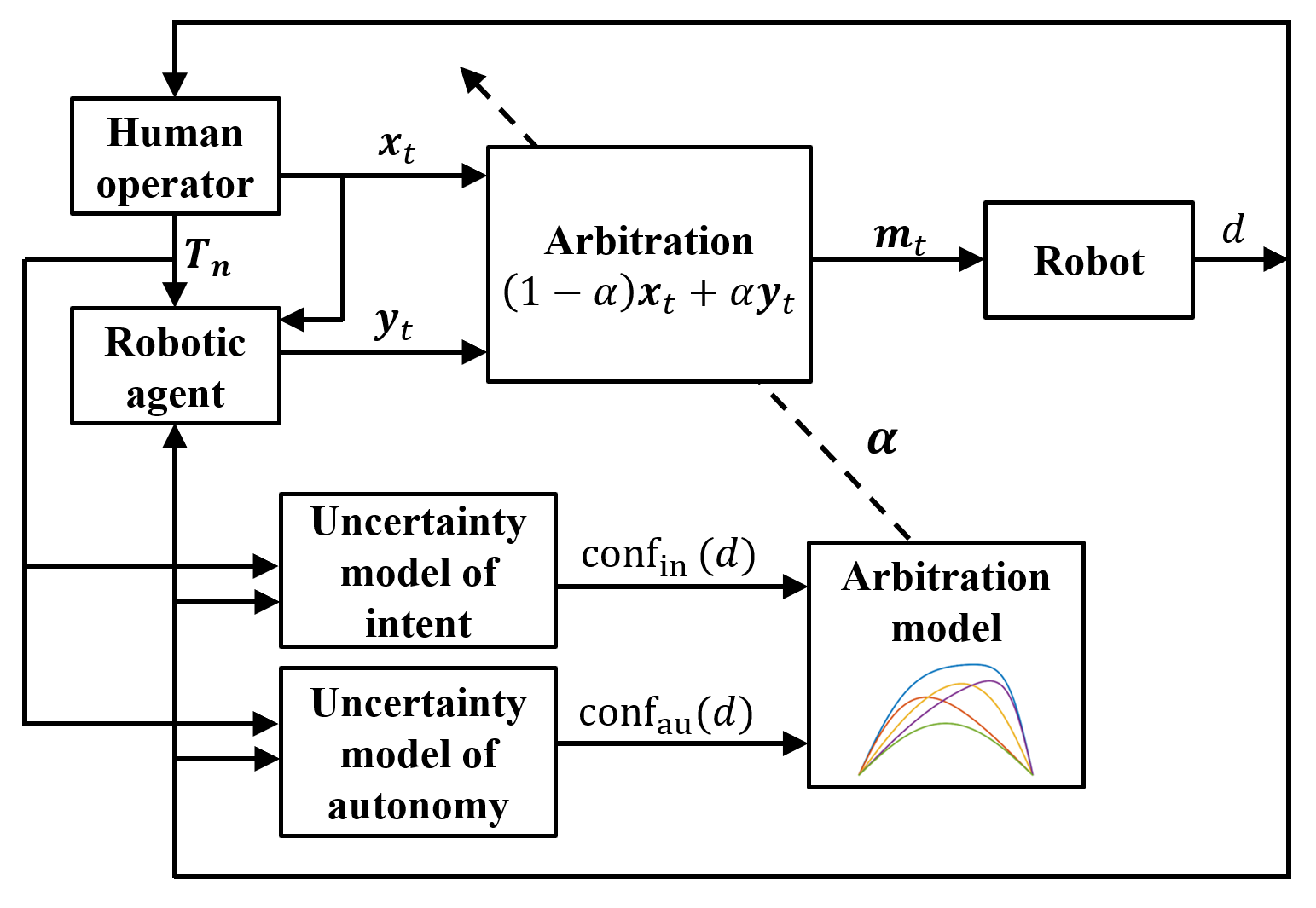}}
\caption{The uncertainty-aware shared control framework.}
\label{fig5}
\end{figure}

According to the above discussion, an uncertainty-aware shared control framework can be developed as shown in Fig. \ref{fig5}. The multimodal intent inference module will infer the human's intended target by observing human's eye-hand movements. The robot trajectory planning module will generate the assistive motion plans based on the inferred target and the robot's current position. The autonomy uncertainty module will evaluate the system's confidence in independently handling the task based on the distance from the current robot position to the target position and the level of autonomy uncertainty. The confidence of assistance will be calculated using $\operatorname{conf}_{\text {hi}}$ and $\operatorname{conf}_{\text {au}}$ as equations \eqref{CIIU6} and \eqref{CAU11}, which will be fed into the control arbitration module for dynamically allocating control power between the human and the autonomous robot in real-time to regulate the robot's action. 

\subsection{Measures of Helpfulness and Friendliness}

\begin{figure}[tb]
\centerline{\includegraphics[width=0.7\linewidth]{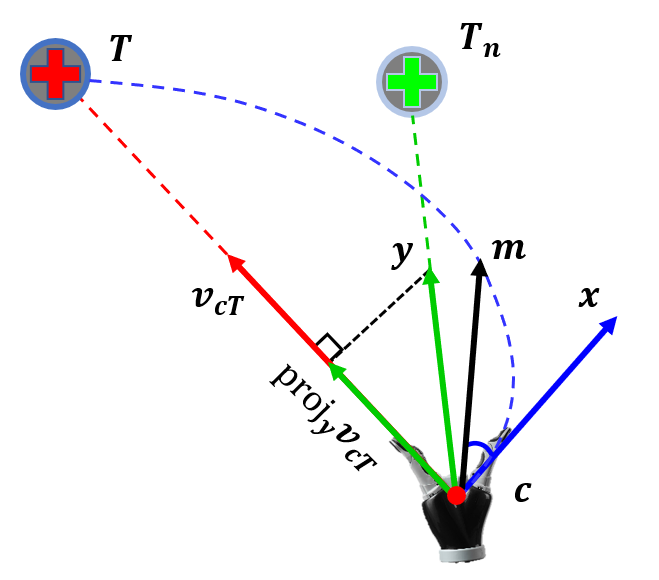}}
\caption{Demonstration of the helpfulness and friendliness definitions. $T$ is the true target, and $T_n$ is the nominal target. $\text{proj}_{y}\boldsymbol{v}_{cT}$ is the projection of $\boldsymbol{y}$ on $\boldsymbol{v}_{cT}$.  }
\label{fig6}
\end{figure}

Fig. \ref{fig6} demonstrates the definition of a robotic agent's helpfulness and friendliness. Both the helpfulness and friendliness are defined at each timestep, and the average values through a complete trial could be calculated as an overall measure. At a certain timestep $t$, the robot's end effector is at $c$, the true target is $T$, and the robotic agent considers $T_n$ as the nominal target due to a combination of the intent uncertainty and autonomy uncertainty. The human operator attempts to approach the true target with a curved trajectory, and the instant motion input of the operator is $\boldsymbol{x}_t= \overrightarrow{\boldsymbol{x}_t}\cdot|\boldsymbol{x}_t|$, where $\overrightarrow{\boldsymbol{x}_t}$ is a unit vector indicates the motion direction and $|\boldsymbol{x}_t|$ is the motion magnitude. In contrast, the robotic agent attempts to approach the nominal target straightly, and its instant motion input is $\boldsymbol{y}_t= \overrightarrow{\boldsymbol{y}_t}\cdot|\boldsymbol{y}_t|$, where $\overrightarrow{\boldsymbol{y}_t}$ is a unit vector indicates the motion direction and $|\boldsymbol{y}_t|$ is the motion magnitude. $\boldsymbol{v}_{cT}$ is the vector points to the target from the current end-effector position, and it can be represented as $\boldsymbol{v}_{cT}= \overrightarrow{\boldsymbol{v}_{cT}}\cdot|\boldsymbol{v}_{cT}|$, where $\overrightarrow{\boldsymbol{v}_{cT}}$ is a unit direction vector pointing to  the  target and $|\boldsymbol{v}_{cT}|$ is the distance to the target.  

The helpfulness $H_t$, of the robotic agent is defined as the $\alpha$-weighted projection of $\overrightarrow{\boldsymbol{y}_t}$ onto $\overrightarrow{\boldsymbol{v}_{cT}}$, which is the weighted unit travel distance in the direction of $\overrightarrow{\boldsymbol{v}_{cT}}$ while traveling $\overrightarrow{\boldsymbol{y}_t}$ \eqref{MHF1}. $H_t$ ranges from -1 to 1, where a positive measure means the end-effector moves closer to the true target with the motion assistance of the robotic agent, while a negative measure indicates the robotic agent interferes the approaching to the target. In extreme conditions when either $|\boldsymbol{y}_t|$ or  $|\boldsymbol{v}_{cT}|$ is zero the robotic agent's helpfulness is defined as zero since the end effector is not moving closer or further from the target. 

The friendliness $F_t$, of the robotic agent is defined as the agreement between the human operator's motion input and the final motion command to the end effector \eqref{MHF2}. It is the cosine of the intersection angle of the $\boldsymbol{x}_t$ and $\boldsymbol{m}_t$.  $F_t$ ranges from -1 to 1. Measure 1 indicates the final motion is along with the human input, and the robotic agent compromises completely and is friendly to the human operator. In contrast, measure -1 indicates the final motion command is in the reversed direction from the operator's input, where the robotic agent is arbitrary and unfriendly. When either $|\boldsymbol{x}_t|$ or $|\boldsymbol{m}_t|$ is zero, it is defined that the robotic agent has a friendliness -1. In addition, when both $|\boldsymbol{x}_t|$ and $|\boldsymbol{m}_t|$ are zeros, the agent's friendliness is 1. 

\begin{equation}
H_{t} = 
\begin{cases}
\alpha \cdot \frac{\boldsymbol{y}_{t} \cdot \boldsymbol{v}_{cT}}{|\boldsymbol{y}_{t}| \cdot |\boldsymbol{v}_{cT}|} & \text{if $|\boldsymbol{y}_{t}| \neq 0$ and $|y_{cT}| \neq 0$} \\
0 & \text{others}
\end{cases}
\label{MHF1}
\end{equation}

\begin{equation}
F_t =
\begin{cases}
\frac{\boldsymbol{x}_{t} \cdot \boldsymbol{m}_{t}}{|\boldsymbol{x}_{t}| \cdot |\boldsymbol{m}_{t}|} & \text{if $|\boldsymbol{x}_{t}| \neq 0$ and $|\boldsymbol{m}_{t}| \neq 0$} \\
1 & \text{if $|\boldsymbol{x}_{t}| = 0$ and $|\boldsymbol{m}_{t}| = 0$} \\
-1 & \text{others}
\end{cases}
\label{MHF2}
\end{equation}

\begin{figure*}[h]
	\centering
	\subfloat[]
	{
		\includegraphics[width = 0.14\linewidth]{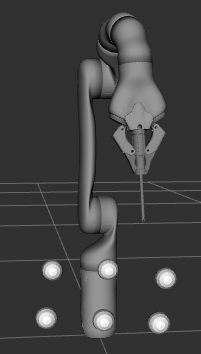}
		\label{fig7a}
	}
	\subfloat[]
	{
		\includegraphics[width = 0.23\linewidth]{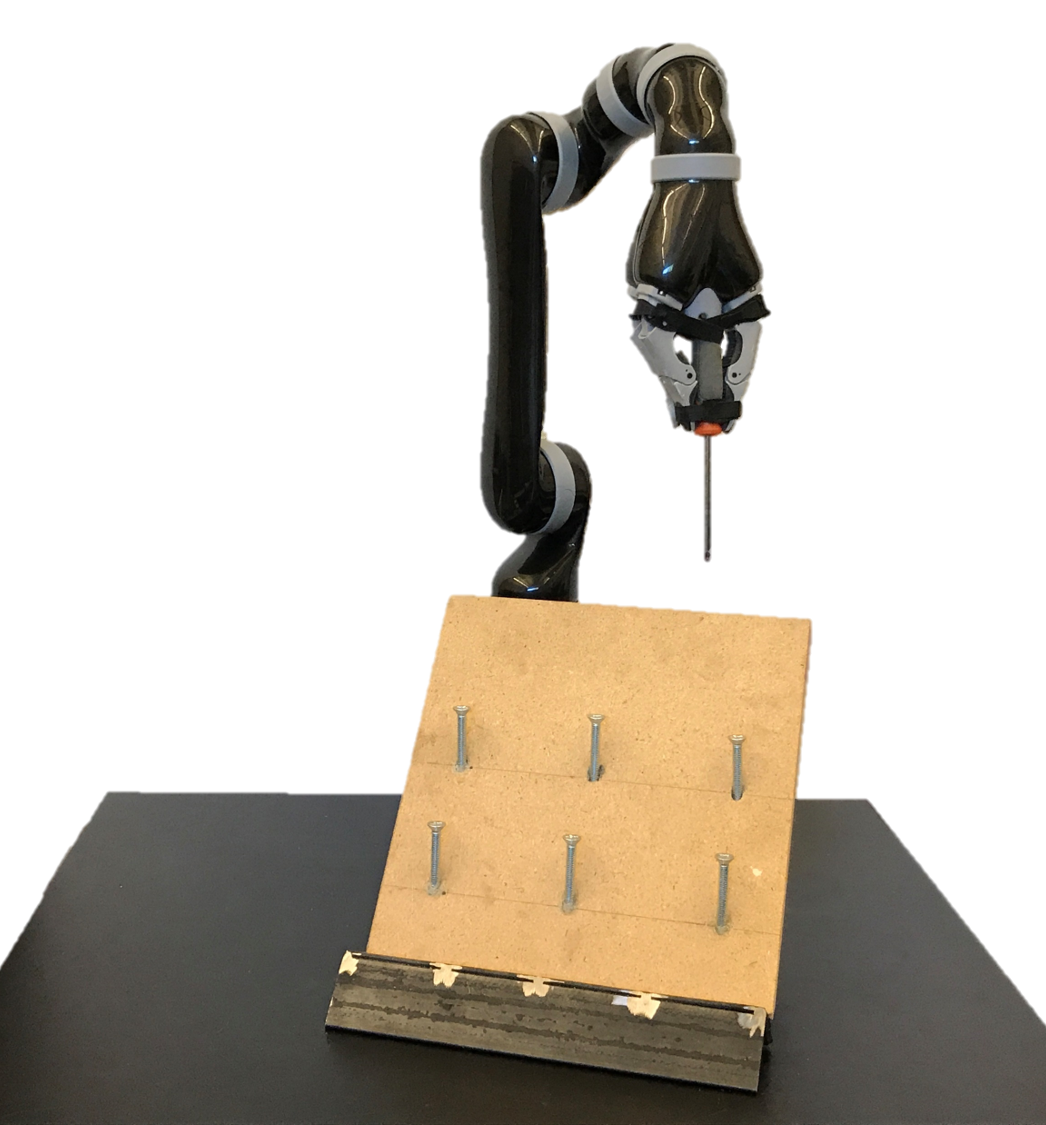}
		\label{fig7b}
	}
	\subfloat[]
	{
		\includegraphics[width = 0.31\linewidth]{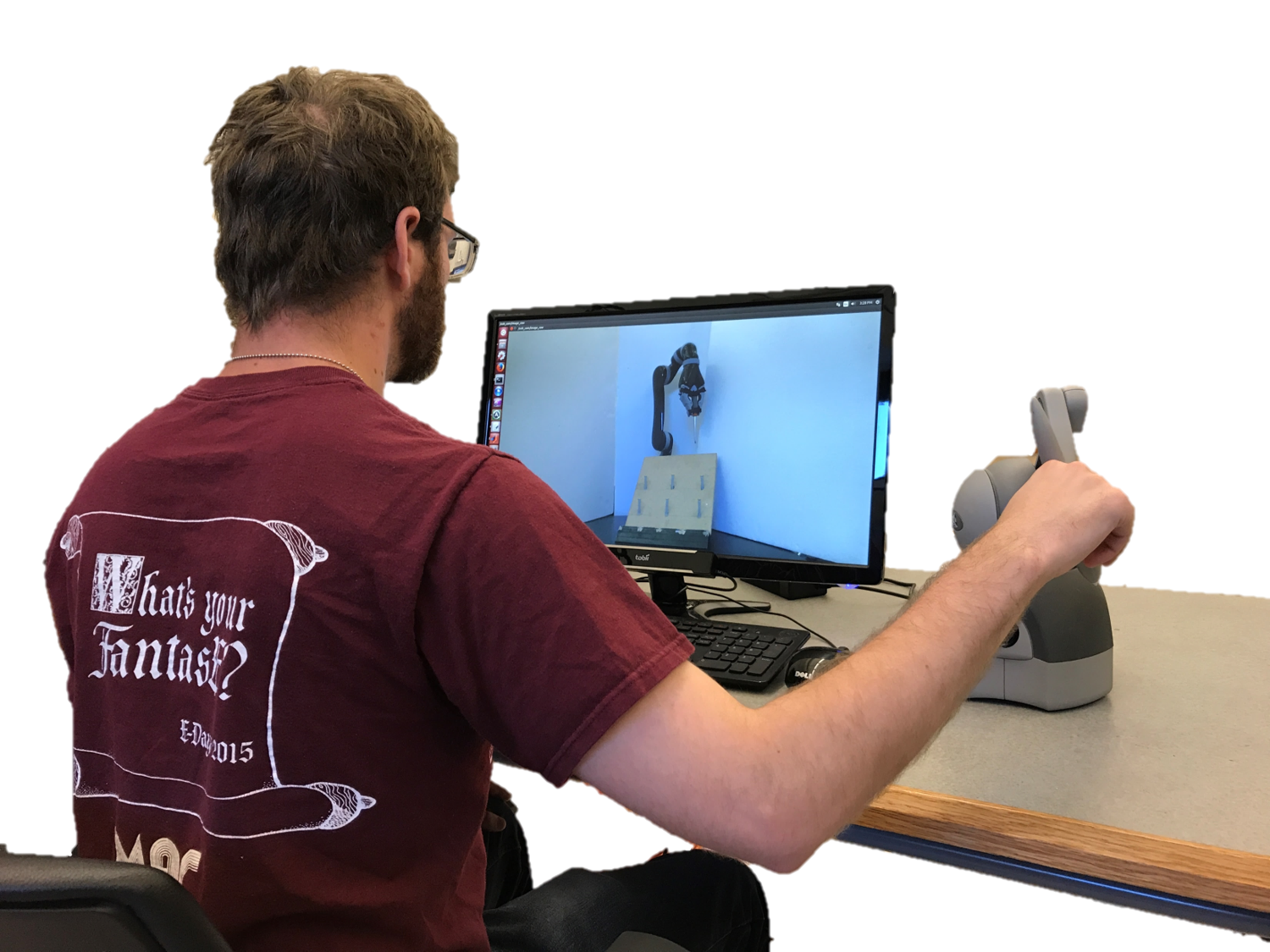}
		\label{fig7c}	
	}
	\caption{Demonstrations of the testing environments. (a) The simulation environment with a virtual MICO robotic arm and six bolts. (b) The robot setup with six bolts for the experiments. (c) The control interface for the experiments.}
	\label{fig7}
\end{figure*}

\section{Simulations and Experiments}

This uncertainty-aware arbitration model is validated within both simulations and real experiments. The tasks are the same in both testing setups where one user teleoperates a MICO robotic arm with a screwdriver in its hand to approach six bolts. The testing setups can be demonstrated with Fig. \ref{fig7}. This setup mimics a manufacturing scenario with teleoperated robots. Six bolts are arranged in two parallel rows with two different heights, and the rows are rotated to avoid the row direction is along the axis direction of either the camera's or robot's frame. The home position of the robotic arm and the relative positions of the bolts to the arms are the same in both setups to achieve comparable results. For comparisons, one traditional positive policy and negative policy were implemented and tested too.

\subsection{Simulation Setup}

In the simulation, the human input was synthesized to have a curved approaching trajectory to the target. The simulated human input at each time step $t$ was synthesized as $\boldsymbol{x}'_t$ in \eqref{SS1}-\eqref{SS2}, and the prime symbol indicates it was a synthesized human input. $\boldsymbol{x}'_t$'s magnitude was defined by a constant $A'$, and its direction was obtained by rotating the direction vector, $\overrightarrow{\boldsymbol{u}_t}$ that pointed to the target from the current robot location, an angle $\theta_t$. $R(\theta_t)$ was the rotation matrix. $\theta_t$ gradually reduced proportionally to the robot-to-target distance, $b_t$, and the initial angle, $\Theta$, was initialized randomly following a normal distribution, $\mathcal{N}(\mu_\theta,\sigma_\theta^2)$. In the simulation, $\mu_\theta$ and $\sigma_\theta$ took values of 20$^\circ$ and 10$^\circ$ respectively. In contrast to the curved trajectory, the robot agent attempted to go straight to the nominal target with its motion input $\boldsymbol{y}_t$ \eqref{SS3}. The robot input's magnitude was equal to the smaller item among the human input magnitude and distance to the nominal target, $d_t$. The nominal target deviated from the target because of human intent inference error and robot perception error (the two uncertainties). It was assumed that the input from the human had a constant speed, and the simulation was run on 20 Hz. 

\begin{equation}
\boldsymbol{x}'_t=R(\theta_t)\overrightarrow{\boldsymbol{u}_t}\cdot A' \label{SS1}
\end{equation}
\begin{equation}
\theta_t=\Theta\cdot\dfrac{b_t}{D}, \Theta \sim \mathcal{N}(\mu_\theta,\sigma_\theta^2) \label{SS2}
\end{equation}
\begin{equation}
\boldsymbol{y}_t=v_t\cdot B; B=\{A',d_t\} \label{SS3}
\end{equation}

Both the human intent inference uncertainty and robot autonomy uncertainty were simulated with various levels. For the intent uncertainty, while approaching a bolt target, we assumed there was a certain time period that the robot considered another bolt as the approaching target and generated motion assistance toward the wrong target. After that certain time, the robot realized this mistake by re-inferring the intent with newly observed eye-hand data and started to assist in approaching the correct target. Thus, the effect of the intent uncertainty can be represented as a time period that the robot treats another object as the wrong target. Six levels of intent uncertainty were simulated, which reflected by the length of the time period of approaching the wrong target. This time period ranged from 0 seconds to 10 seconds. For the autonomy uncertainty, a certain offset was added to the actual location of the bolts in a randomized direction. Even though the offset's direction was random, the magnitude was the same. Six levels of autonomy uncertainty were simulated ranging from 0 cm to 10 cm. The intent uncertainty only affected the robotic agent in the defined short time period, while the autonomy uncertainty affected the robotic agent all the time. Six levels of intent uncertainty and six levels of autonomy uncertainty gave 36 uncertainty combinations. 

Three arbitration policies, the proposed arbitration model, and traditional positive and negative policies were tested with the same initial condition under the 36 uncertainty combinations. In one simulation trial, the human operator and the robotic agent cooperated to approach the six bolts as six independent approaching tasks. The inputs from the human operator and the robotic agent were blended using one of the three policies. Three simulation trials with applying each arbitration policy once comprised a set of simulation trials. The human initial offset angle, $\Theta$, was initialized first in a simulation set so that three policies were tested with the same initial condition for comparison fairness. Approaching one bolt was successful if the robot's end-effector was close enough to the target bolt's head, while it was a failure when the end effector was stuck at the nominal target. The success rate and the completion time of successful runs were recorded. The friendliness and the helpfulness were calculated at each time step during the approaching process, and the mean value was used to summarize the whole approaching process. One hundred simulations sets were performed for statistical analysis, and it resulted in 600 approaching tasks for each policy under an uncertainty combination.

\subsection{Experiment Setup}

In the experiment, the human user sent the control command through a Geomagic Touch haptic joystick to approach three of the six bolts as one approaching trial. The robotic agent's input still pointed to the nominal target and took a magnitude of the smaller value of the human input or the distance to the nominal target. To reduce the difficulty of the task, the approaching was only being performed in the camera's X-Y plane and left the depth control free. It reduces the task difficulty as previous research concluded that the control in the depth direction was most difficult in teleoperation. For creating an identical testing environment across all participants and three arbitration policies, the intent inference uncertainty was simulated in the same way as in the simulation but with two uncertainty levels of 5 seconds and 10 seconds. The autonomy uncertainty was simulated in the same way as in the simulation with two levels of 1 cm and 3 cm offsets. 

Each participant performed one approaching trial with each arbitration policy under one test setting. In the meantime, the testing order was randomized to minimize the order effects. The success rate and completion time were recorded. The inputs of the operator and robotic agent were recorded to compute the friendliness and helpfulness. After performing each trial, a short questionnaire was provided to obtain subjects' positive and negative opinions on the assistance provided by the robotic agent. The questionnaire consisted of eight assessment statements about the robotic agent's performance, and the subjects needed to mark their agreement level to each statement. 

\section{Results}

\subsection{Results of the Simulations}

Four measures were taken from the tests in simulations: the task completion time, task success rate, and robotic agent's helpfulness and friendliness. The task completion time is counted only for the successful trials, but the helpfulness and friendliness were computed for all trials. These three measures (completion time, helpfulness, and friendliness) are presented with boxplots to display the distribution. The exact values of completion time, friendliness, and helpfulness are not displayed in the plots as they are subjected to the pre-set human input, and the changing trend of those measures among various uncertainty conditions is more meaningful. Similarly, the exact uncertainty settings are abstracted to six uncertainty levels, Level 0 (L0) to Level 5 (L5). Level 0 means there is no uncertainty, and level 5 is the highest uncertainty. The Mann-Whitney U test was performed on the measures to statistically compare the positive and negative policies to the bell-shaped policy. Two significance levels were taken with p $<$ 0.001 as high significance (solid dots) and p $<$ 0.01 as moderate significance (circles). No statistical significance is notated with a cross.

\subsubsection{Success Rate and Completion Time}

\begin{figure}[t]
\centerline{\includegraphics[width=0.85\linewidth]{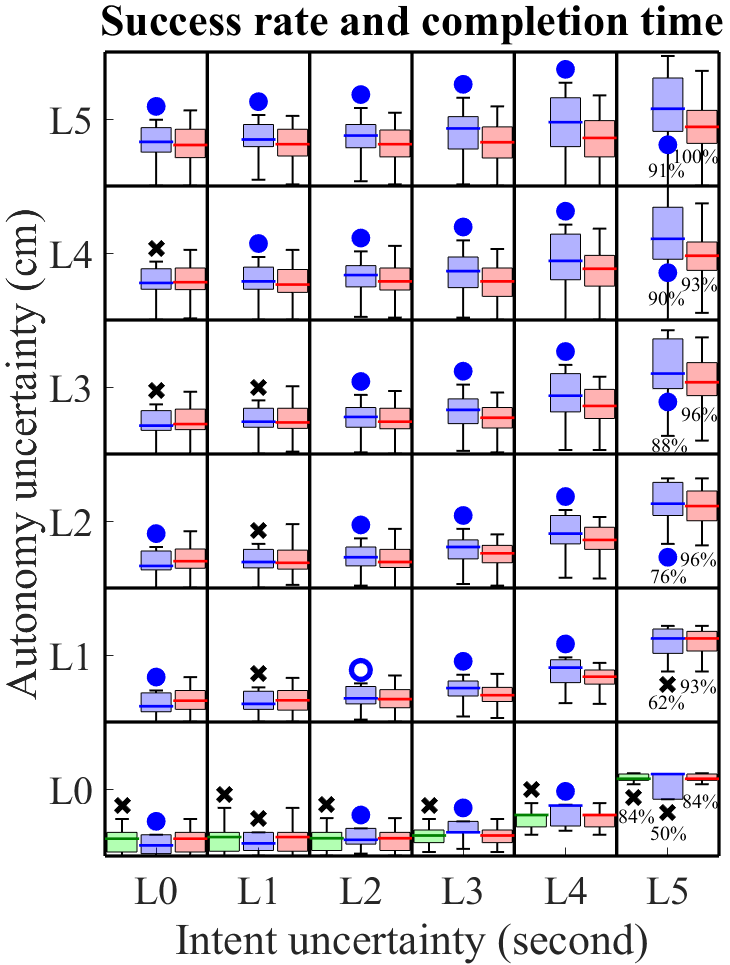}}
\caption{Task completion time (boxplots) and task success rate (percentages) when applying the positive policy (green), the negative policy (blue), and the bell-shaped policy (red) in simulations. Comparing to the bell-shaped policy, solid dots indicate the completion time is different with statistical significance (p $<$ 0.001), circles indicate the completion time is different with a moderate significance (p $<$ 0.01), and black cross marks indicate the difference is not statistically significant.}
\label{fig8}
\end{figure}

Fig. \ref{fig8} summarizes the task completion times and task success rates of each policy under a certain uncertainty. If a policy achieved a success rate of less than 20\%, its completion time distribution is discarded as it had insufficient data to perform the analysis and discussions, and if using a policy achieved a success rate of 100\% this success rate is not displayed to keep the plot concise. Other than these, the success rates are displayed as percentages under the distribution plots. 

The success rates of applying three arbitration policies greatly varied with uncertainty conditions, however, the bell-shaped policy had the best success rate all conditions. When no autonomy uncertainty was present and the intent uncertainty was not high (less than L5), three policies achieved a success rate of 100\%. The negative and bell-shaped policies continued to have a success rate of 100\% when the autonomy uncertainty increased from L1 to L5. However, the positive policy resulted in the robot's end-effector stuck at the nominal target due to the autonomy uncertainty and had success rates lower than 20\%. The L5 intent uncertainty undermined all three policies' success rates, but the bell-shaped policy was consistently more successful than the other policies. Moreover, at L5 intent uncertainty, the success rate of applying the negative and bell-shaped policies increased with the increase of the autonomy uncertainty. This could because of the mutual effects on two types of uncertainties.

The completion times of applying three policies increased gradually when increasing the uncertainty levels. The bell-shaped policy was more efficient in overall. The negative policy and the bell-shaped policy are mainly compared here, as the positive policy either had the same completion time or its success rate was too low when autonomy uncertainty was absent or present. When both types of uncertainty were low, using the negative policy could accomplish the task quicker, and their difference was statistically significant. However, the negative policy's advantage becomes smaller with increases of uncertainty in either intent or autonomy. The completion time of using the negative policy became comparable with the bell-shaped policy and eventually overpassed its. In 22 out of 36 uncertainty conditions, the bell-shaped policy's completion was shorter. It was also noted that both policies' completion time had larger increases when the intent uncertainty rose from L3 to L4 and from L4 to L5.

\subsubsection{Helpfulness}

\begin{figure}[t]
	\centerline{\includegraphics[width=0.85\linewidth]{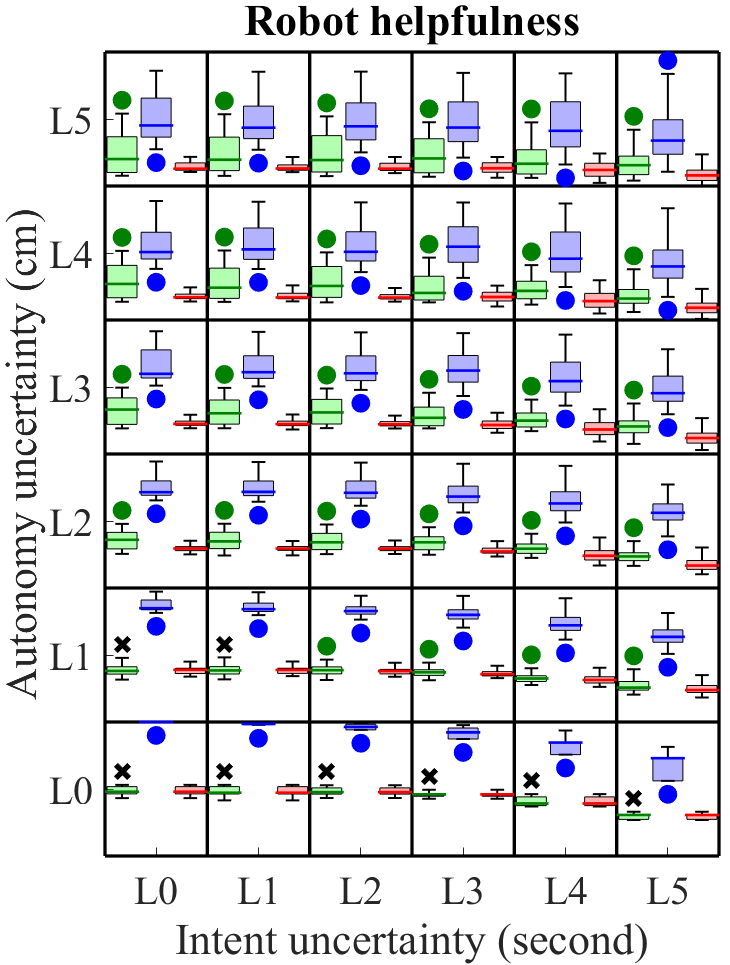}}
	\caption{The robotic agent's helpfulness during the approaching when applying the positive policy (green), the negative policy (blue), and the bell-shaped policy (red) in simulations. Comparing to the bell-shaped policy, solid dots indicate a policy's helpfulness is significantly different (p $<$ 0.001), circles indicate the helpfulness is different with a moderate significance (p $<$ 0.01), and black cross marks indicate the difference is not statistically significant.}
	\label{fig9}
\end{figure}

Fig. \ref{fig9} summarizes the helpfulness of the robotic agent in various testing conditions with three arbitration policies. Generally, the negative policy had the highest helpfulness across all uncertainty conditions, and the bell-shaped policy had the lowest helpfulness in most of the conditions. 

When no autonomy uncertainty was presented, the positive policy and the bell-shaped policy had the same helpfulness, which was significantly lower than the negative policy. While adding the autonomy uncertainty, the helpfulness of the three policies became lower. In the meantime, the difference between the negative policy and the bell-shaped policy was reducing, and their distribution overlaps appeared and grew larger. When both uncertainties were low (less than L2), the helpfulness of bell-shaped and positive policies were not statistically different. However, the helpfulness of the bell-shaped policy became lower than the positive policy when increasing the uncertainties. It is also noticed that the autonomy uncertainty greatly increased the distribution variance of the helpfulness.

\begin{figure}[t]
	\centerline{\includegraphics[width=0.85\linewidth]{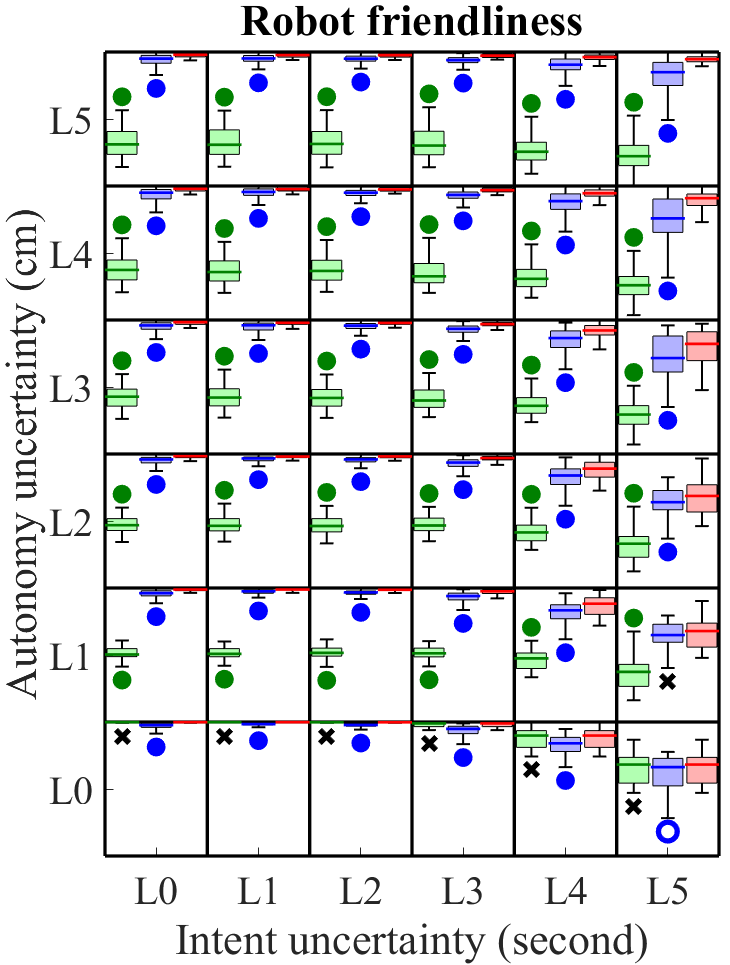}}
	\caption{
	The robotic agent's friendliness during the approaching when applying the positive policy (green), the negative policy (blue), and the bell-shaped policy (red) in simulations. Comparing to the bell-shaped policy, solid dots indicate a policy's friendliness is significantly different (p $<$ 0.001), circles indicate the friendliness is different with a moderate significance (p $<$ 0.01), and black cross marks indicate the difference is not statistically significant.}
	\label{fig10}
\end{figure}

\subsubsection{Friendliness}

Fig. \ref{fig10} summarizes the friendliness of the robotic agent in various uncertainty conditions with three arbitration policies. Among all the uncertainty conditions, the bell-shaped policy had the highest friendliness, and its friendliness was statistically different from others. When the uncertainties were low, the friendliness advantage of the bell-shaped policy was small but grew larger when the intent uncertainty was high.

When no autonomy uncertainty was presented, the positive policy and the bell-shaped policy had the same friendliness. Both positive and bell-shaped policies' friendliness was slightly higher than the negative policy, and this difference was statistically significant. Presence of the autonomy uncertainty could cause great drops to the positive policy's friendliness, however, continuously increasing the autonomy uncertainty only led to mild decreases. Moreover, the friendliness decreases of the negative and bell-shaped policies were hardly observable when only increasing the autonomy uncertainty at lower levels of intent uncertainty (L4 or lower). This made the positive policy significantly unfriendly compared to the other policies. It was also noted that three policies' friendliness had larger declines when the intent uncertainty rose from L3 to L4 and from L4 to L5.

\begin{figure}[t]
	\centerline{\includegraphics[width=\linewidth]{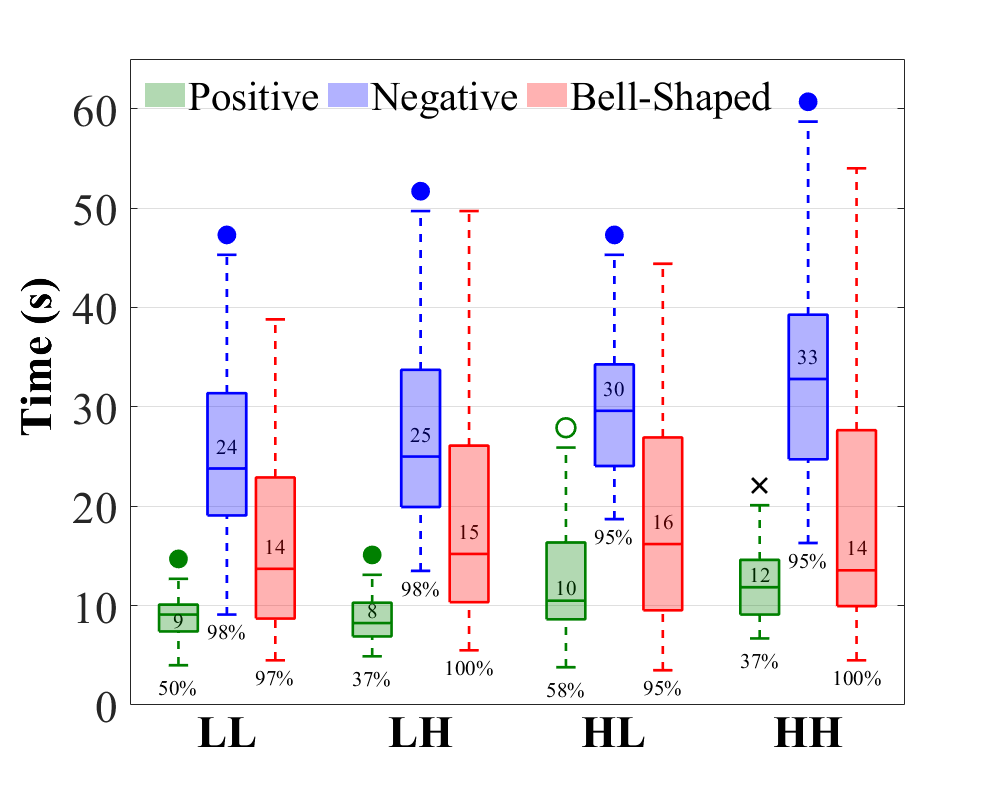}}
	\caption{Task completion time (boxplots) and task success rate (percentages) when applying the positive policy (green), the negative policy (blue), and the bell-shaped policy (red) in the experiments. Comparing to the bell-shaped policy, solid dots indicate a policy's completion time is significantly different (p $<$ 0.01), circles indicate the completion time is different with a moderate significance (p $<$ 0.05), and black cross marks indicate the difference is not statistically significant. From left to right, there are test settings of low intent, low autonomy uncertainty (LL), low intent, high autonomy uncertainty (LH), high intent, low autonomy uncertainty (HL), and high intent, high autonomy uncertainty (HH).}
	\label{fig11}
\end{figure}

\subsection{Results of the Experiments}

Data from 370 approaching trials were collected from the experiment testing from ten participants. Four objective measures and one subjective questionnaire were recorded from the experiment. Four testing conditions are notated by the combination of uncertainty levels (such as LL indicates low intent and autonomy uncertainty, and LH indicates low intent but high autonomy uncertainty). The Mann-Whitney U test was also performed on the measures to separately compare the positive and negative policies to the bell-shaped policy. Two significance levels were selected with p $<$ 0.01 as high significance (solid dots) and p $<$ 0.05 as moderate significance (circles).

\subsubsection{Success Rate and Completion Time}

Fig. \ref{fig11} summarizes the task completion times and task success rates of the experiments. The success rates are displayed as percentages under the corresponding boxplot, and the time median is displayed inside the box.

It is evident that the bell-shaped policy is the most effective arbitration policy among all uncertainty conditions. Overall, the bell-shaped policy had the highest success rate, which was greatly higher than the positive policy and was higher by 6\% than the negative policy on average. In contrast, the positive policy failed a lot during the experiment, and in two settings, it had a success rate that was low as 37\%. Even though the successful trials of using the positive policy had a very low completion time, it was still sufficiently reasonable to conclude the positive policy function worst in the experiment. The success rate of the negative policy was close to the bell-shaped policy, but its completion time was much longer than the bell-shaped policy, and their difference was statistically significant. 

The bell-shaped policy stably functioned well in four uncertainty conditions, as its accuracy and completion time had small variances. It suggested that the bell-shaped policy was robust to the uncertainty variances. In contrast, both positive and negative policies were affected by the uncertainty changes. The positive policy had a lower success rate when the autonomy uncertainty was high, while the negative policy was sensitive to the increase of intent uncertainty. 

\begin{figure}[t]
	\centerline{\includegraphics[width=\linewidth]{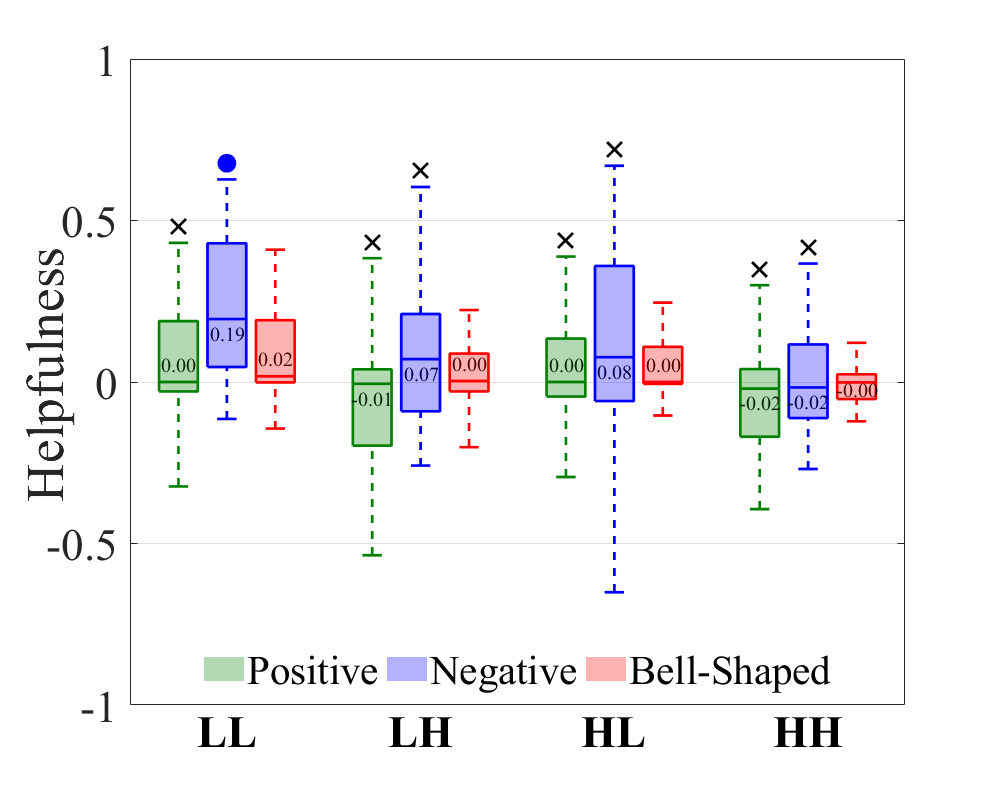}}
	\caption{The robotic agent's helpfulness during the approaching when applying the positive policy (green), the negative policy (blue), and the bell-shaped policy (red) in experiments. Solid dots indicate the data is significantly different (p $<$ 0.01), and black cross marks indicate the difference is not statistically significant. }
	\label{fig12}
\end{figure}

\subsubsection{Helpfulness}

Fig. \ref{fig12} summarizes the helpfulness of the robotic agent in the experiments when three arbitration policies were applied in various uncertainty conditions. It shows that the negative policy had the highest helpfulness in three uncertainty conditions, however, their differences were not statistically significant in most uncertainty conditions. The helpfulness of the positive and bell-shaped policies was close and not statistically different. 

When comparing the three policies' helpfulness distributions in various conditions (the portion between the 25th percentile and the 75th percentile), the bell-shaped policy's helpfulness was relatively stable, while the negative policy varied the most. It was also noted that the negative policy's helpfulness was decreasing when increasing either type of uncertainty, and reached its lowest when both types of uncertainty were high. In addition, the data suggested that the positive policy's helpfulness was higher when the autonomy uncertainty was low.

\begin{figure}[t]
	\centerline{\includegraphics[width=\linewidth]{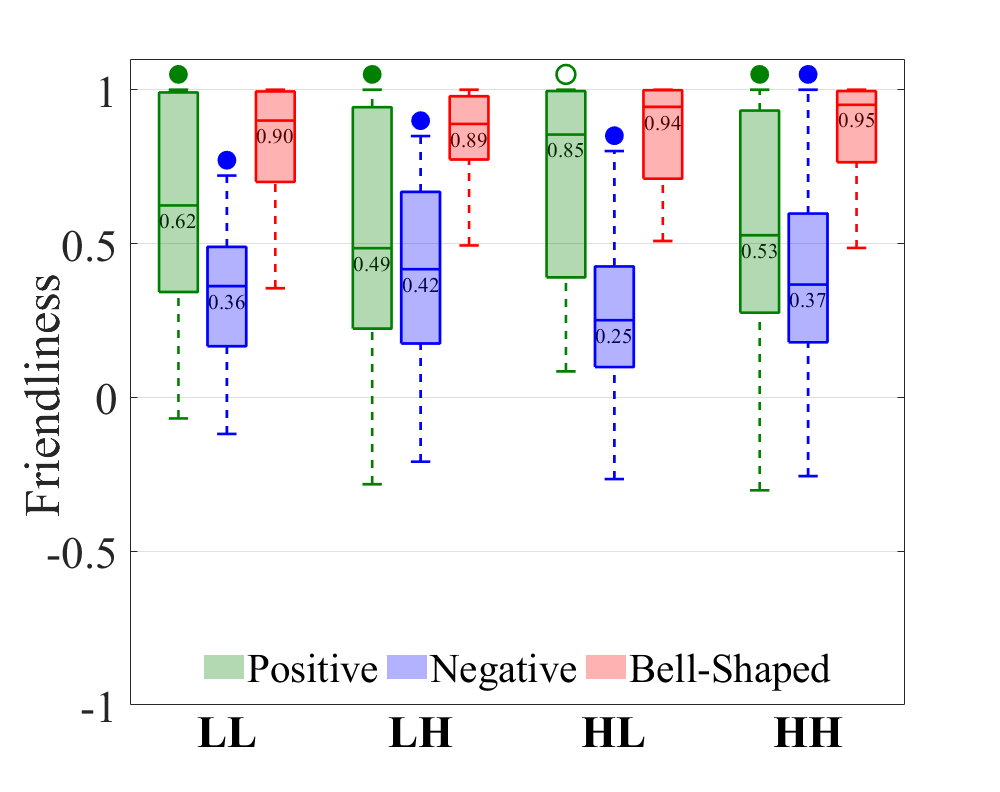}}
	\caption{The robotic agent's friendliness during the approaching when applying the positive policy (green), the negative policy (blue), and the bell-shaped policy (red) in experiments. Solid dots indicate the data is significantly different (p $<$ 0.01), and circles indicate the data is different with a moderate significance (p $<$ 0.05). }
	\label{fig13}
\end{figure}

\begin{figure}[t]
	\centerline{\includegraphics[width=\linewidth]{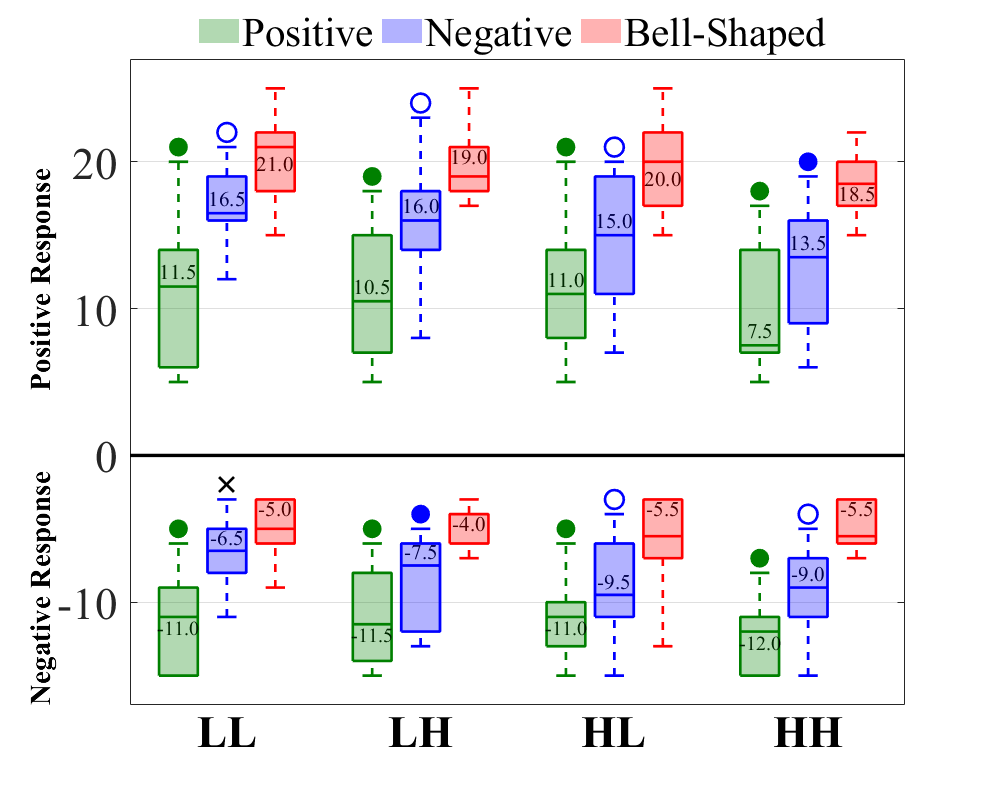}}
	\caption{Participants' subjective evaluation of the positive policy (green), the negative policy (blue), and the bell-shaped policy (red) in experiments. Solid dots indicate the data is significantly different (p $<$ 0.01), circles indicate the data is different with a moderate significance (p $<$ 0.05), and black cross marks indicate the difference is not statistically significant.}
	\label{fig14}
\end{figure}

\subsubsection{Friendliness}

Fig. \ref{fig13} summarizes the friendliness of the robotic agent in the experiments when three arbitration policies were applied in various uncertainty conditions. It shows that the bell-shaped policy was the most friendly across all uncertainty conditions, and its friendliness advantage is apparent and statistically significant. In contrast, the negative policy was the most unfriendly.

The distribution of the bell-shaped policy's friendliness did not change much with varying uncertainty conditions, which suggested its robustness in the term of friendliness. The positive policy's friendliness decreased when autonomy uncertainty was increased. Interestingly, it is noted that there was a friendliness increase for the positive policy when intent uncertainty was higher, and there was a small friendliness increase for the negative policy when the autonomy uncertainty was higher.

\subsubsection{Questionnaires}

Fig. \ref{fig14} summarizes the subjective evaluation of the robotic agent when three arbitration policies were applied in various uncertainty conditions. The accumulative scores from positive and negative assessment portions are plotted separately, and the higher the score is, the better the subjective assessment is. It is apparent that the bell-shaped policy had the highest scores, and its score distribution was moderately different from the negative policy and significantly different from the positive policy. In addition, the positive policy had the lowest score in all the uncertainty conditions. 

\begin{figure*}[t]
	\centering
	\subfloat[]
	{
		\includegraphics[width = 0.235\linewidth]{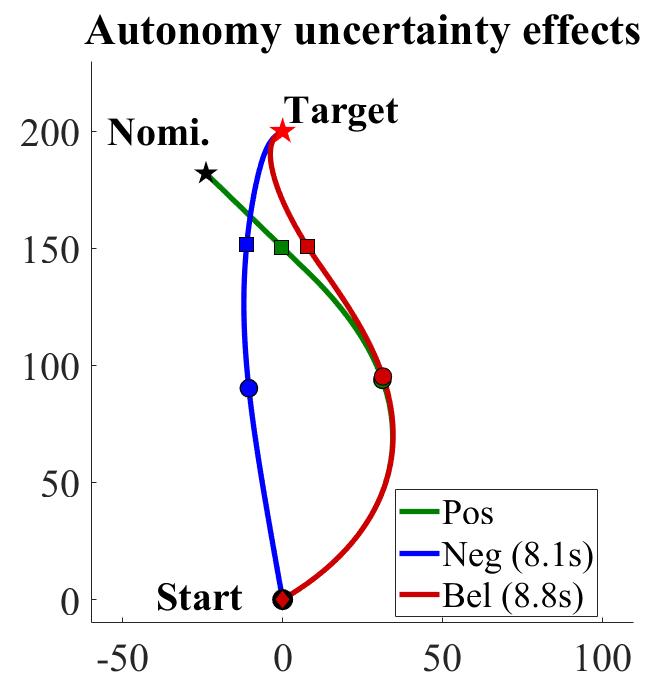}
		\label{PIa}
	}
	\subfloat[]
	{
		\includegraphics[width = 0.235\linewidth]{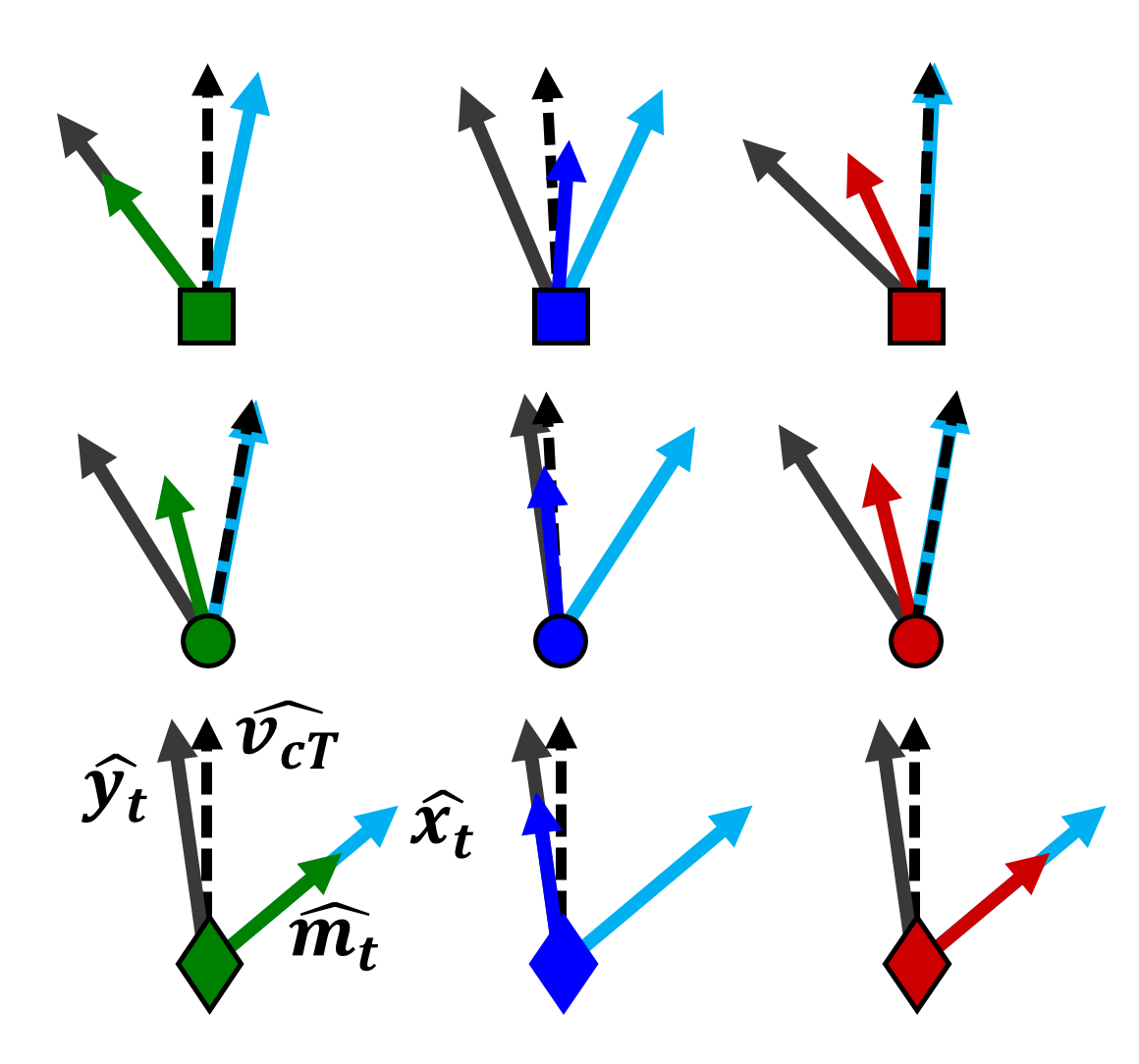}
		\label{PIb}
	}
	\subfloat[]
	{
		\includegraphics[width = 0.235\linewidth]{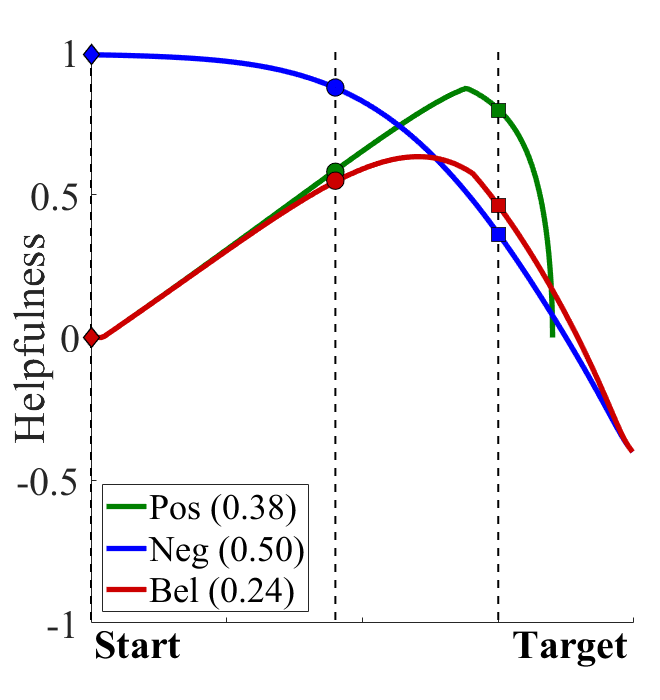}
		\label{PIc}	
	}
	\subfloat[]
	{
		\includegraphics[width = 0.235\linewidth]{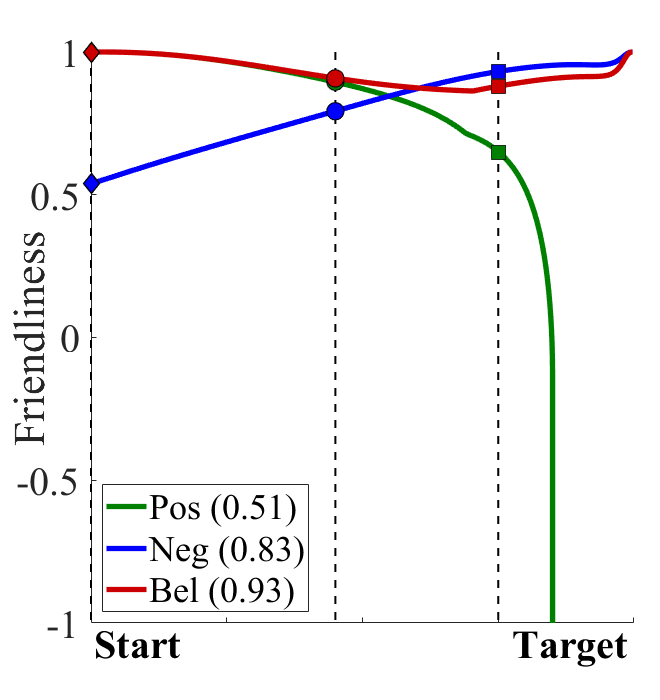}
		\label{PId}	
	}
	\caption{(a) The trajectories of the robot's end-effector when using three arbitration policies. Applying the positive policy resulted in approaching failure, and the negative and bell-shaped policies used 8.1s and 8.8s to complete the approaching respectively. (b) Nine points were taken from the three trajectories when the end effector was respectively at the starting location (diamond markers), 110 mm away (circle markers), and 50 mm away (square markers) from the target. When the end effector was at these locations the directions of the human motion ($\boldsymbol{x}_t$), the robot motion ($\boldsymbol{y}_t$), the final motion input to end effector ($\boldsymbol{m}_t$), and the direction pointing to the target from the current position of the end effector ($\boldsymbol{v}_{cT}$) show in different colors. While the end effector approaches the target, the robotic agent's helpfulness and friendliness are plotted in (c) and (d). }
	\label{PI}
\end{figure*}

The score distributions of the bell-shaped policy in various uncertainty conditions were similar, which suggested that the bell-shaped policy functioned consistently in various uncertainty conditions and had the participants consistently given positive evaluations. The positive policy had consistently low scores in three uncertainty conditions, and its score in the condition of high intent and high autonomy uncertainties was the lowest. It seems that the negative policy was evaluated higher when the autonomy uncertainty was low.

\section{Discussion}

\subsection{Policy Illustration}

Fig. \ref{PIa}-\ref{PId} demonstrate how each arbitration policy worked and affected the motion of the end-effector using a mini 2D simulation. The robot's end-effector moved on a plane where the true target, nominal target, and the end effector were on. The end effector started from the origin to approach a target 200 mm away on the Y-axis. The human inputs were synthesized to have a curved trajectory as the robotic agent in the simulation tests. This mini simulation only considered the autonomy uncertainty, which was simulated the same way as in the simulations and experiments, and the uncertainty was at a moderate level to show distinguish differences between various policies. The trajectories of the end effector as a result of the blended human and robotic agent inputs show in Fig. \ref{PIa}, and three arbitration policies resulted in different motion trajectories. When using the positive policy, the approaching task failed with the end effector stuck at the nominal target. In contrast, applying the negative and bell-shaped policies successfully accomplished the task, and the negative policy (8.1s) had a shorter completion time than the bell-shaped policy (8.8s). Along three trajectories, nine points were taken when the end effector was respectively at the starting location (diamond markers), 110 mm away (circle markers), and 50 mm away (square markers) from the target. When the end effector was at those locations the directions of the human input motion ($\overrightarrow{\boldsymbol{x}_t}$), the robot input motion ($\overrightarrow{\boldsymbol{y}_t}$), the final motion input to end effector ($\overrightarrow{\boldsymbol{m}_t}$), and the direction pointing to the target from the current position of the end effector ($\overrightarrow{\boldsymbol{v}_{cT}}$) show in Fig. \ref{PIb} with various colors. Fig. \ref{PIc} and \ref{PId} plot the robotic agent's helpfulness and friendliness respectively, and the average helpfulness and friendliness are shown in the legends.

When the end effort was at the start location, the motion inputs of the human and robotic agents were the same across three policies. However, due to the various arbitration policies, different arbitration weights were assigned to the robotic agent, and final motions to the end effector were different. The positive and bell-shaped policies both assigned 0 to the robotic agent; thus the final motion command was along the human input. Consequently, the robotic agent was completely friendly to the human operator as it compromised completely and provided no help. In contrast, the negative policy had the final motion command along the robotic agent's input, which resulted in helpfulness near 1 as the intersection angle between $\overrightarrow{\boldsymbol{y}_t}$ and $\overrightarrow{\boldsymbol{v}_{cT}}$ was near 0. However, due to the large offset between the $\overrightarrow{\boldsymbol{x}_t}$ and $\overrightarrow{\boldsymbol{m}_t}$, the robotic agent's friendliness was around 0.5. 

While approaching the target, the arbitration weights varied with the changed spatial relationship between the target and end effector and resulted in varying helpfulness and friendliness. From the start point to the point 110 mm away, more control power was assigned to the robotic agent by the positive and bell-shaped policy. Consequently, the robotic agent became more helpful as it was dragging the end effector close to the target even toward the nominal target. In the meantime, the end effector became less following the human operator, and friendliness gradually reduced. In contrast, the robotic agent's helpfulness with the negative policy was reducing, but its friendliness was increasing due to the smaller arbitration weight to the robotic agent and more compromise to the human operator. 

When the end effector was 50 mm away from the target, it followed the robotic agent completely to the nominal target due to the assertion of the positive policy. As drifting from the target largely, the helpfulness and friendliness of the robotic agent both started to decrease rapidly. In contrast, the negative and bell-shaped policies had returned a majority of the control power to the human operator at this close range. The robotic agent's helpfulness was reducing because its arbitration weight was reduced and the motion to the nominal target had less contribution to moving close to the target at this close range. In addition, the robotic agent became more friendly as it compromised more. 

When the end effector was far from the target, moving toward the nominal target also greatly contributed to the approaching of the target. Due to this reason, the negative policy had the shortest completion time and higher helpfulness. However, the introduction of the intent uncertainty could weaken this advantage or even convert it to a negative policy's disadvantage since it could drag the end effector to somewhere else.

\subsection{Policy Evaluation}

Three arbitration policies are comprehensively evaluated by combining their results in simulations and experiments. Firstly, using the bell-shaped policy had the best success rate over the other policies across the various uncertainty conditions, and it also resulted in shorter completion time. Using the negative policy could achieve a comparable but lower success rate than using the bell-shaped policy, and the negative policy's completion time was much longer. The positive policy resulted in the worst success rates as it was so sensitive to the autonomy uncertainty. Secondly, while using the bell-shaped policy, the robotic agent was less helpful yet more friendly. In contrast, the robotic agent provided the most help but was less friendly using the negative policy. When using the positive policy, the robotic agent's helpfulness was lower than using the negative policy and comparable with using the bell-shaped policy, and its friendliness was lower than using the bell-shaped policy. These two measures together indicated a robotic agent's intrusiveness of helping the human operator. Thirdly, the bell-shaped had the highest subjective scores followed by the negative policy then the positive policy. This subjective score was the operator's comprehensive evaluation of the combination of success rate, completion time, helpfulness, and friendliness. 

In summary, the bell-shaped policy is evident to be more effectively arbitrate control power between the human operator and the robotic agent for better cooperation and better performance. Using the bell-shaped policy achieved higher success rates and shorter completion time regardless of the uncertainty variances in both simulations and experiments. It demonstrated the robustness of the bell-shaped policy in coping with the system uncertainty. Moreover, the bell-shaped policy is subjectively preferred by the human operators. The bell-shaped policy had lower helpfulness and higher friendliness, and this indicates the bell-shaped policy regulated the robotic agent's help to be more nonintrusive but effective.

\subsection{Friendliness and Helpfulness}

The newly developed friendliness and helpfulness were validated to quantitatively and objectively evaluate how an assistive robotic agent was friendly to the human operator and how effective in helping the human operator to accomplish the task. The robotic agent's friendliness and helpfulness in simulations and experiments have many matched characters, which could be strong evidence that the metric definitions were reasonable and valid. Compared to the existing success rate and task completion time as the overall performance measurements (defined as the macro level in this paper as it is to quantify overall performance), friendliness and helpfulness could be microscopical or macroscopic which measure a robotic agent's behavior at each timestep or through an overall trial. Moreover, these two metrics are cooperation-orientated that quantify how well two agents' cooperation instead of the performance-orientated success rate and completion time. Two metrics enable researchers to explicitly explain how an arbitration policy affects the subjective and objective outcome, and thus to provide novel insights into the robotic agent and the shared-control paradigm. 

One example of the new findings revealed is that many researchers believed that it was better for the robot to provide more help in the collaboration, which, however, was proved not always valid by the helpfulness measures from the simulations and experiments. The robotic agent using the negative policy had the highest helpfulness but achieved lower success rates and used longer time to accomplish the tasks than the less helpful bell-shaped policy. In the meantime, the robotic agent using the negative policy was also less friendly. High helpfulness meant the robotic agent was strongly pulling the robot's end effort toward its perceived target; however, it could result in competition for the control of the end effector as the robotic agent was following a different trajectory plan that was unnatural for the human operator. This new finding inspires us to reconsider how a robotic agent should provide its assistance in order to achieve better performance. 

Even though great insights into the robotic agent's assistance were revealed by the helpfulness and friendliness, the definition could be further improved. The current definition did not take consideration of the motion inputs' magnitudes. The input's magnitude was certainly critical, but how it should be considered in an effective manner needs more investigations. Two metrics' values at extreme conditions were defined discretely, and it arose questions on their legibility. In the meantime, both metrics may need to be scaled to increase their separating capability. Currently, both metrics ranges from 0 to 1 mostly as shown in Fig. \ref{PId}. In addition, even the negative policy aggressively dragged the end effector to a different path while the bell-shaped policy had a better match with the operator's desired trajectory, the friendliness difference between the bell-shaped policy and negative policy was often small numerically. For example, the bell-shaped policy's trajectory was much closer to the human's than the negative policy in Fig. \ref{PIa}, but their friendliness difference was only 0.1. Increasing the separating capability can increase various policies' measure distinction and facilitate the comparison.

\subsection{Simulations and Experiments}

The simulations and experiments were complimentary in evaluating the three policies. Simulations enabled more extensive tests of the three policies with less effort to reveal each policy's characters in various uncertainty conditions, and real experiments were essential to verify the simulation results and the policies' practical effectiveness since the great difficulty in simulating a real physical environment and human intuition. Even though the experiments were not at the same scale as the simulations, the high consistency in the simulation results and experiment results gave us high confidence in the effectiveness of the bell-shaped policy and the findings derived from the results. 

There were three minor inconsistent instances in the simulation results and experiment results, which did not undermine the derived conclusions but deserved further investigations. The first one was the positive policy's success rate. In experiments, the positive policy still had the lowest success rates, but they were not that low as in the simulations. The second instance was the statistical difference between each policies' helpfulness. In the simulation, the bell-shaped policy's helpfulness was statistically different from others. However, the experiment results only matched the simulation results in general but did not have the same statistical significance. The third inconsistent instance was that the positive policy's friendliness was higher than the negative one. The human operators' intuition could have greatly contributed to the success of using the positive policy, while they managed to avoid the nominal target and reach the true target. The randomness in human operators' approaching trajectories could have reduced the margins between three policies' helpfulness distributions. The extra-long stuck time before ending the simulation and more standstill moments in experiments were the major reasons why the positive policy had higher friendliness than the negative policy. In the simulation of the positive policy, the human operator would attempt to escape from the stuck condition before calling it a failure, and this attempting time was set longer than necessary. During this stuck time, the friendliness kept measuring -1 according to the definition as the friendliness trail shown in Fig. \ref{PIb}, and this extra-long trail lowered the positive policy's friendliness. However, the experimenter terminated the approaching trial shortly after the end effector stuck. In addition, there were more standstill moments when using the positive policy in experiments. During the standstill moment, the robotic agent's friendliness was continuously measured as 1. Fine-tuning the simulation termination criteria and implementation and refining the friendliness definition are desirable for future investigation. 

There was a common limitation in both simulations and experiments that the task was simplified for the human operator. Operating a robot could be a very difficult task for the human operator due to the disembodiment and physical discrepancy, and this was the main reason why robotic assistance was desirable. However, this difficulty was not simulated in the simulations and had been simplified a great amount in the experiments. It would be necessary to test the bell-shaped and negative policies in an experimental setup closer to the practical application.

\section{Conclusion}

In this paper, we investigated the arbitration relationship between a human operator and a robotic agent in shared-control teleoperation. We believe that the lack of consideration of the multiple types of uncertainty in the human-robot system was one reason for the great inconsistency of the arbitration policies. To fill this gap, we modeled the multi-source uncertainty from the human intent inference process and robotic automation system. Different types of uncertainty affect the control arbitration differently. We then developed an arbitration model that comprehensively fused the uncertainty and regulated the control arbitration. The developed uncertainty model was based on a 3D Gaussian distribution, which was general to easily incorporate more types of uncertainty. Meanwhile, the arbitration model was also general and extendable to incorporate other types of uncertainty. The arbitration model was then evaluated with simulations and experiments with comparisons of the existing arbitration policies. The new arbitration model outperformed or performed equivalently to the current policies in all the uncertainty combinations across all the measures. In addition, we developed helpfulness and friendliness as two new objective and quantitative metrics to reveal how well a robotic agent cooperated with the human operator under an arbitration policy and explain how the policy functioned and influenced the motion commends in dynamic at the micro level. The two new metrics can better analyze the arbitration policies to uncover the limitations or strongpoints. With the work in this paper, we expect the advancement of shared control in teleoperation for practical deployments. 
	
\section{Acknowledgments}
This material is based on work supported by the US NSF under grant 1652454. Any opinions, findings, and conclusions or recommendations expressed in this material are those of the authors and do not necessarily reflect those of the National Science Foundation.

	\bibliographystyle{IEEEtran}
	\bibliography{uncertaintypolicy}
\end{document}